\documentclass{article}

\usepackage{PRIMEarxiv}

\usepackage[utf8]{inputenc} 
\usepackage[T1]{fontenc}    
\usepackage{hyperref}       
\usepackage{url}            
\usepackage{booktabs}       
\usepackage{amsfonts}       
\usepackage{nicefrac}       
\usepackage{microtype}      
\usepackage{lipsum}
\usepackage{fancyhdr}       
\usepackage{graphicx}       
\graphicspath{{media/}}     
\usepackage{xspace}
\usepackage{amsmath,amssymb,amsfonts}
\usepackage{nomencl}

\title{A Survey of Offline and Online Learning-Based Algorithms for Multirotor UAVs
}

\author{
  Serhat S\"onmez, Matthew J. Rutherford, Kimon P. Valavanis \\
  D. F. Ritchie School of Engineering and Computer Science \\
  University of Denver \\
  CO, USA\\
  \texttt{firstname.surname}@du.edu}

\begin{document}
\maketitle

\begin{abstract}
Multirotor UAVs are used for a wide spectrum of civilian and public domain applications. Navigation controllers endowed with different attributes and onboard sensor suites enable multirotor autonomous or semi-autonomous, safe flight, operation, and functionality under nominal and detrimental conditions and external disturbances, even when flying in uncertain and dynamically changing environments. During the last decade, given the faster-than-exponential increase of available computational power, different learning-based algorithms have been derived, implemented, and tested to navigate and control, among other systems, multirotor UAVs. Learning algorithms have been, and are used to derive data-driven based models, to identify parameters, to track objects, to develop navigation controllers, and to learn the environment in which multirotors operate. Learning algorithms combined with model-based control techniques have been proven beneficial when applied to multirotors. This survey summarizes published research since 2015, dividing algorithms, techniques, and methodologies into offline and online learning categories, and then, further classifying them into machine learning, deep learning, and reinforcement learning sub-categories. An integral part and focus of this survey are on online learning algorithms as applied to multirotors with the aim to register the type of learning techniques that are either hard or almost hard real-time implementable, as well as to understand what information is learned, why, and how, and how fast. The outcome of the survey offers a clear understanding of the recent state-of-the-art and of the type and kind of learning-based algorithms that may be implemented, tested, and executed in real-time.
\end{abstract}


\section{Introduction}  \label{sec: intro}

Unmanned aerial vehicles (UAVs) have witnessed unprecedented levels of growth during the last 20 years, with civilian and public domain applications spanning power line inspection \cite{martinez2014towards}, monitoring mining areas \cite{ren2019review}, wildlife monitoring and conservation \cite{olivares2015towards}, border protection \cite{bassoli2019virtualized}, building and infrastructure inspection \cite{carrio2016ubristes}, and precision agriculture \cite{li2016real}, to name but a few applications. Although different UAV types and configurations have been utilized for such applications, multirotor UAVs, particularly quadrotors, are the most commonly and widely used, due to their perceived advantages, effectiveness, hovering capabilities, and efficiency during flight.  

A plethora of conventional and advanced model-based linear, linearized, and nonlinear controllers have already been derived and used for multirotor navigation and control. However, recently, learning-based algorithms and techniques have gained momentum because they facilitate and allow, among other things, for: i.) data-driven system modeling that may also include model updates; ii.) combined data-driven and model-based modeling and control and parameter identification; iii.) data-driven parameter identification; iv.) data-driven environment model development, and v.) pure learning-based control. Stated different, learning-based approaches add to the model-based formulation, they enhance multirotor modeling and control, and they offer alternatives to learning, modeling and understanding the environment.  

Learning-based algorithms are basically divided into offline and online learning, although there exist some learning algorithms that include an offline and an online learning component. Most researchers have extensively studied and applied to different families of (linear and nonlinear) systems offline learning-based algorithms. Derivation and implementation of online learning-based algorithms to multirotor UAVs is a relatively recent area of research that has attracted increased interest because of the real-time implementability potential, which may facilitate continuous anytime online learning. This momentum has motivated the present survey.  

To begin with, a review of the literature reveals that there exists considerable published research addressing the use of learning algorithms for UAV control. The emphasis of already published surveys is on developing and adopting machine learning, deep learning, or reinforcement learning algorithms. To be specific, Carrio et al.~\cite{carrio2017review} center around deep learning methods that are applied to UAVs, while in Polydoros and Nalpanditis~\cite{polydoros2017survey} emphasis is in model-based reinforcement learning techniques applied to robotics, but also with applications to UAVs. Machine learning algorithms for UAV autonomous control are explored by Choi and Cha \cite{choi2019unmanned}, while Azar et al.~\cite{azar2021drone} investigate deep reinforcement learning algorithms as applied to drones. Most recently, Brunke et al.~\cite{brunke2022safe} presented several learning algorithm-based applications in robotics, including multirotor UAVs. 

The common theme of already published surveys is that they discuss different offline learning-based control algorithms that may be, or have been applied to different UAV types, but they are not real-time implementable. Therefore, in contrast to the existing surveys, the focus of this research is on also registering the online learning-based algorithms that have shown potential, and/or have been implemented and tested on multirotor UAVs.  

Without loss of generality, for the purpose of this survey, the below provided 'attributes' are considered important and they facilitate the review process. The list is by no means complete, nor unique; it may be modified and enhanced accordingly based on set objectives. Note that in what follows, the terms 'agent' and 'multirotor' are used interchangeably.  

\begin{enumerate}
  \item \textit{Navigation task:} This refers to the (autonomous or semi-autonomous) function the multirotor needs to accomplish, given a specific controller design and/or configuration. 
  \item \textit{Learning:} This refers to 'what' the agent learns in order to complete the navigation task.
  \item \textit{Learning Algorithm:} This refers to the specific algorithm that needs to be followed for the agent to learn. Inherent in this attribute is 'what' is being learned by the agent, and 'how'. 
  \item \textit{Real-time applicability:} This refers to 'how fast' learning is achieved, and 'how computationally expensive' is the learning algorithm, which, basically dictates whether learning is applicable in hard real-time, or, in almost hard real-time. Stated differently, the answer to 'how fast' determines the implementability of the learning algorithm. Calculation of the algorithm's computational complexity may also provide additional information on 'how fast' the agent learns. 
  \item \textit{Pros \& Cons:} This refers to the advantages and limitations of the underlying learning approach, which, in unison with all other attributes, determines the overall applicability and implementability of the learning approach on multirotor UAVs.
 \end{enumerate}

The rest of the survey is organized as follows: Section~\ref{sec: background information} provides background information and related definitions, which is deemed essential for clarification and classification purposes. Section~\ref{sec: offline learning} summarizes offline learning, and provides a detailed Table reflecting published research, also stating what is being learned. The review of offline learning techniques is essential for completeness purposes, and to also understand the differences between offline and online learning. Section~\ref{sec: online learning} dives into online learning approaches. An overview of each learning method is presented, along with what is being learned and why, advantages, and disadvantages, and obtained results. Discussion and conclusions are offered in Section~\ref{sec: discussion and conclusion}.

\section{Background Information} \label{sec: background information}

Key concepts and definitions are presented next. Related and relevant literature is cited to support statements, findings, and observations. This information is adopted and used throughout the paper; it also helps to correctly classify reviewed learning-based algorithms, when needed.

\subsection{Definitions:} \label{subsec: definition}

\begin{figure}[ht]
\begin{center}
\includegraphics[width=10.5 cm]{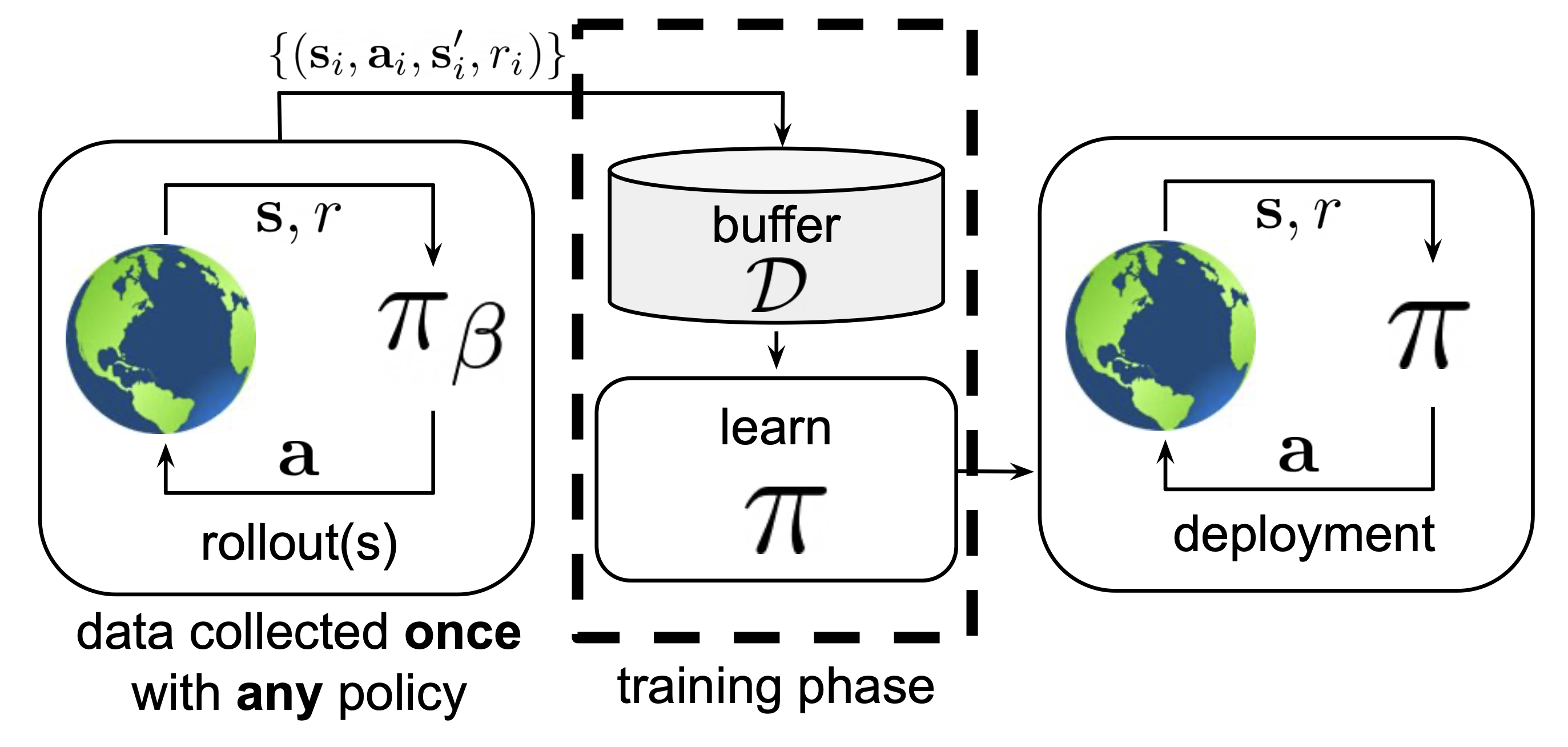}
\caption{Offline reinforcement learning block diagram illustration \cite{levine2020offline}.}
\label{offlineRL}
\end{center}
\end{figure}

\begin{figure}[ht]
\begin{center}
\includegraphics[width=10.5 cm]{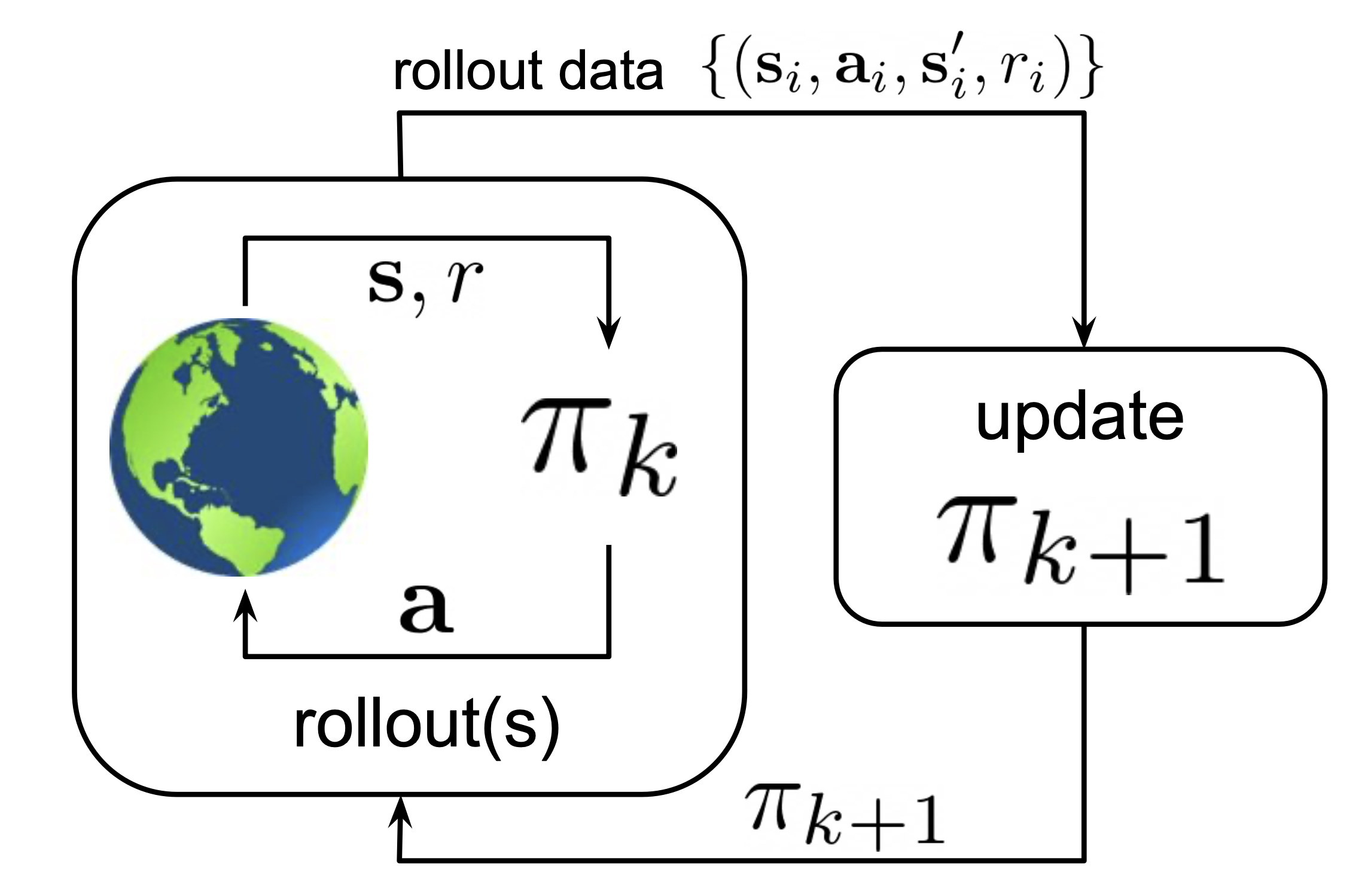}
\caption{Online reinforcement learning block diagram illustration \cite{levine2020offline}.}
\label{OnlineRL}
\end{center}
\end{figure}

\textbf{\textit{Reinforcement Learning:}} Reinforcement Learning, RL, is a machine learning technique in which an agent communicates with the environment to find the best action using the state space and a reward function. RL includes four main components; a policy, a reward signal, a value function, and optionally, an environment model \cite{carrio2017review}, \cite{sutton2018reinforcement}. RL may be either online or offline. The general configuration of the offline RL and the online RL approaches are shown in Figures~\ref{offlineRL} and \ref{OnlineRL}.

\textbf{\textit{Policy:}} In the context of learning algorithms, a policy determines how the learning agent behaves at a given moment. A policy maps the discerned states of the environment to the actions that the agent should take when in those states \cite{sutton2018reinforcement}. The policy may reflect either on-policy or off-policy options.

\textbf{\textit{On-policy:}} On-policy techniques relate to a policy that is used to make decisions. Policy Iteration, Value Iteration, Monte Carlo for On-Policy, and State-Action-Reward-State-Action (SARSA) algorithms are representative examples of on-policy algorithms  \cite{sutton2018reinforcement}. 

\begin{figure}[ht]
\begin{center}
\includegraphics[width=10.5 cm]{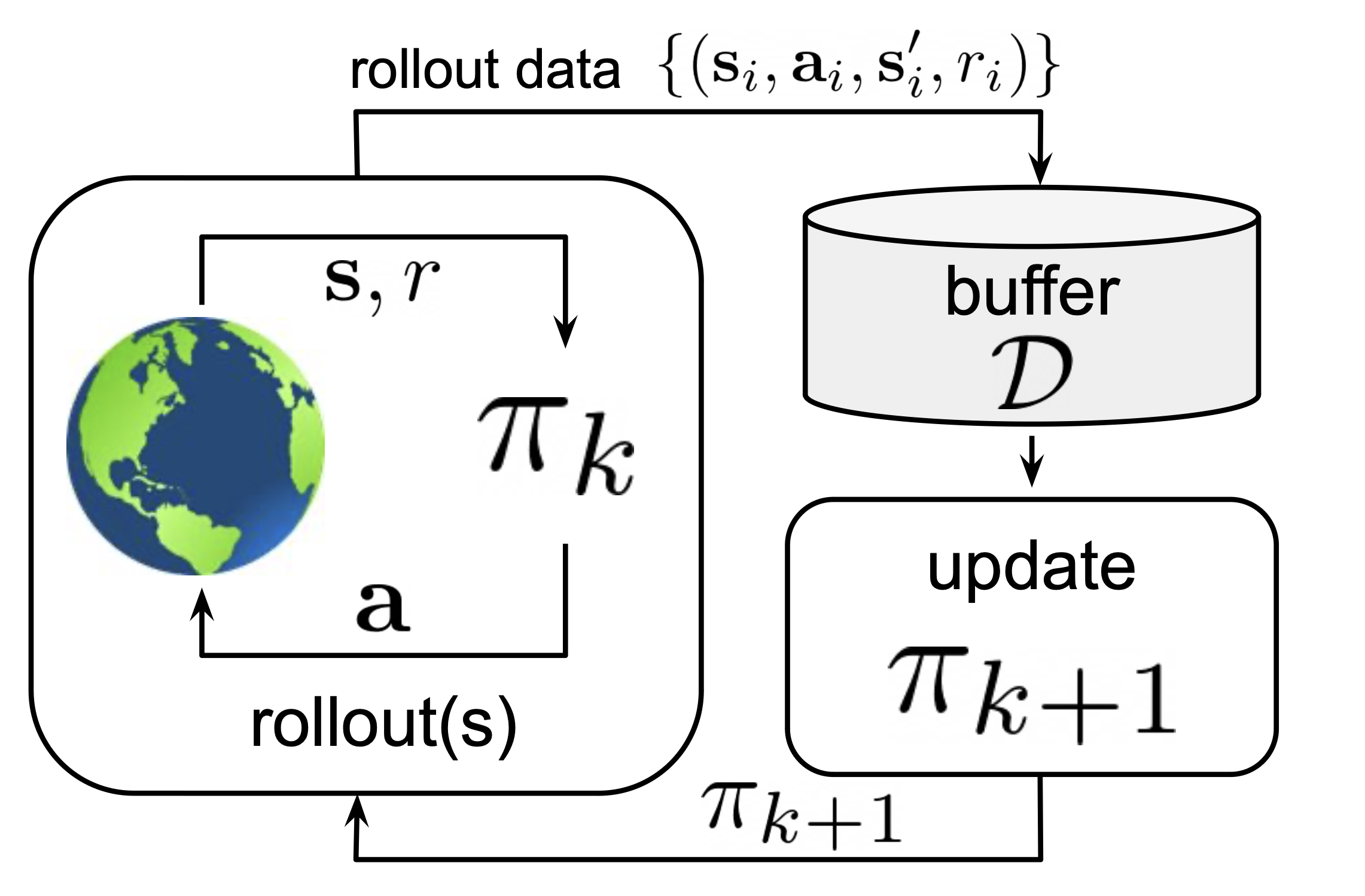}
\caption{Off-policy RL algorithm configuration \cite{levine2020offline}.}
\label{off-policy RL}    
\end{center}
\end{figure}

\textbf{\textit{Off-policy:}} Off-policy methods refer to a policy that is dissimilar to that used to produce data. Q-learning and Deep Deterministic Policy Gradient (DDPG) are examples of off-policy algorithms \cite{sutton2018reinforcement}, see Figure~\ref{off-policy RL} for the configuration of the off-policy RL algorithm. The agent experience is the input to a data buffer $\mathcal{D}$, which is also called a replay buffer. Each new policy $\pi_{k+1}$ is trained by utilizing the samples of all previous policies $\pi_0$, $\pi_1$, $\ldots$ , $\pi_k$ that are stored in $\mathcal{D}$.

On-policy methods are generally less complicated than off-policy ones, and they are contemplated first. Since data is produced by a different policy in off-policy methods, they converge more slowly. Regardless, off-policy methods provide more powerful and general alternatives \cite{sutton2018reinforcement}. 

\textbf{\textit{Reward signal:}} A reward signal is a target in a RL algorithm. The agent receives a single number from the environment in each time step, the reward. The only objective of the agent is to maximize the total reward over time. The reward signal provides an immediate sense of the current state and indicates what events are beneficial or detrimental to the agent \cite{sutton2018reinforcement}. 

\textbf{\textit{Value Function:}} A value function determines what events are advantageous to the agent, long-term. 

\textbf{\textit{Environment Model:}} An environment model helps mimic and replicate the behavior of the environment. It allows for making inferences about how the environment will behave. 

In what follows, the definitions and fundamental concepts of offline, online, supervised and unsupervised learning, machine and deep learning, are detailed.

\textbf{\textit{Offline Learning:}} In offline learning, the learning algorithm trains an agent or artificial neural network (ANN). The agent interacts with the environment and it is updated during the training process. However, the agent is never updated after training is completed. Bartak and Vykovsky~\cite{bartak2015any} and Edhah et al.~\cite{edhah2019deep} present representative examples of offline machine learning (ML) and deep learning (DL) algorithms, respectively. The studies of Xu et al.~\cite{xu2018monocular}, Rodriguez at al.~\cite{rodriguez2019deep}, and Yoo et al.~\cite{yoo2020hybrid} are representative examples of offline RL algorithms. Observing the configuration diagram of offline RL in Figure 1, a dataset ($\mathcal{D}$) is collected by the behavior policy (${\pi}_{\beta}$) with the help of the states (\textbf{\textit{s}}) and the reward function (\textbf{\textit{r}}). A policy ($\pi$) is trained by using $\mathcal{D}$. The training process does not interact with the Markov Decision Process (MDP). The trained policy ($\pi$) is deployed to control the system. The policy ($\pi$) interacts with the environment using the states (\textbf{\textit{s}}) and the reward function (\textbf{\textit{r}}), and creates the action space (\textbf{a}). 

\textbf{\textit{Online Learning:}} In online learning, according to Hoi et al.~\cite{hoi2021online}, the learner keeps on learning and improves prediction to reach the best possible one by interacting with the environment as shown in Figure 2. A policy ($\pi_k$) creates the action space to interact with the environment using the states (\textbf{\textit{s}}) and the reward function (\textbf{\textit{r}}). Then, $\pi_k$ is updated by using the roll-out data including states, actions, future states, and the reward function. After the updated policy ($\pi_{k+1}$) is determined, it is replaced with the current policy ($\pi_k$).

Note that for the purposes of this survey, the definition of online learning is extended to account for cases where the agent continues the learning process during operation, in real-time, even after completing the offline learning process, or without any basic learning. This extension allows for 'anytime learning' while the underlying system continues to function, and accounts for the ability to modify the system model and also adapt its parameter values (time-varying systems).

\textbf{\textit{Supervised Learning:}} In supervised learning, the learning algorithm learns from the classified and labeled dataset \cite{carrio2017review}.

\textbf{\textit{Unsupervised Learning:}} In unsupervised learning, the learning algorithm utilizes unlabeled data, which are collected from sensors, to learn the proposed task. Unsupervised learning techniques are widely used in RL \cite{carrio2017review}.

\textbf{\textit{Machine Learning:}} Machine learning, ML, is a component of Artificial Intelligence (AI), in which tasks are learned (or imitation of tasks is learned) from collected data \cite{carrio2017review}.  

\textbf{\textit{Deep Learning:}} Deep learning, DL, is a category or subset of ML that involves the use of deep neural networks, DNNs, with input, hidden, and output layers to model and solve complex problems.  

Moreover, RL techniques and methods are divided into model-based and model-free ones. 

\textbf{\textit{Model-based:}} In a model-based method, the agent predicts the future states and reward and also chooses the action that provides the highest expected reward \cite{sutton2018reinforcement}. Models and planning are used to solve RL problems. 

\textbf{\textit{Model-free:}} In a model-free method, the agent does not utilize the environment model but makes decisions by only using trial-and-error approaches \cite{sutton2018reinforcement}.

The main difference between model-based and model-free methods is that the former relies on planning, while the latter relies on learning \cite{sutton2018reinforcement}.

A Google Scholar Search since 2015 returns the paper distribution that is illustrated in Figure 4, which shows the number of offline and online learning published papers that deal with multirotor UAVs. 

\begin{figure}
\begin{center}
\includegraphics[width=13.5cm]{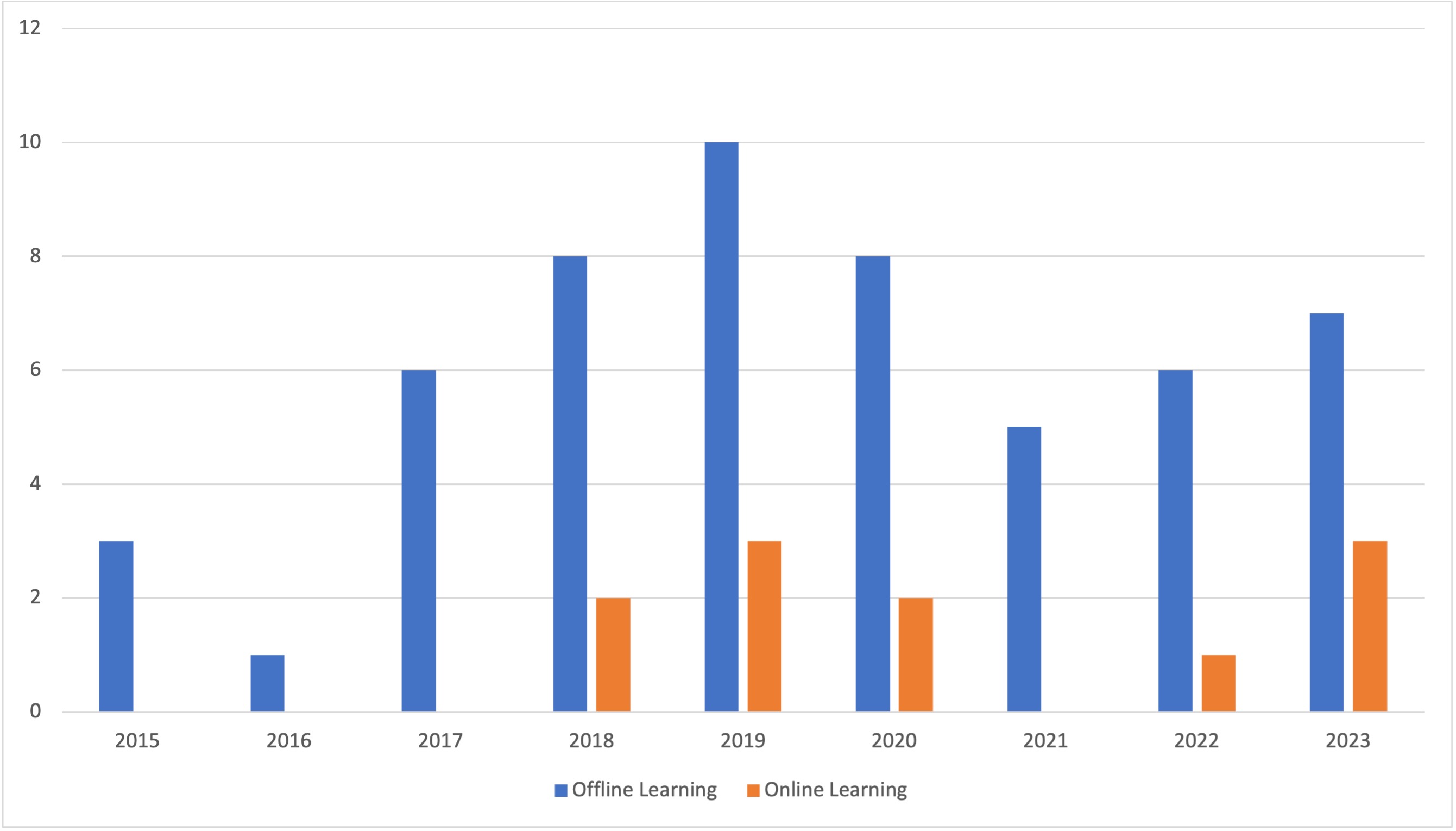}
\caption{Publications of online and offline learning algorithms for control of multirotor UAVs since 2015 based on Google Scholar search.}
\label{article_number}
\end{center}
\end{figure}

The next Section reviews offline learning techniques.

\section{Offline learning} \label{sec: offline learning}

In offline learning, the system may be trained either using collected and/or provided data (supervised learning), or, alternatively, by using feedback before its actual operation, without using any data (unsupervised learning). In this case, when operating in real-time, the agent, or the neural network (NN), is not updated nor affected by the environment. 

Table~\ref{Tab: Offline learning} summarizes offline learning and RL approaches that have been applied to multirotor UAVs. The Table includes the publication year of the paper, the adopted or derived learning model, the application task/domain, as well as what is being learned. 55 articles have been reviewed and classified either as machine learning, ML, or deep learning, DL, or reinforcement learning, RL, approaches (as indicated by the authors). In cases where no information is provided in the reviewed papers, the classification follows the provided definitions in Section~\ref{sec: background information}. 

\begin{table*}[]
\caption{Offline Learning Papers}
\label{Tab: Offline learning}
\resizebox{\textwidth}{!}{
\begin{tabular}{|c|c|c|c|l|}
\hline
\textbf{Year} & \textbf{Authors}                                     & \textbf{Learning Model} & \textbf{Application - Task}                 & \textbf{What is being Learned}                                         \\ \hline \hline
2015          & Bartak et al. \cite{bartak2015any}                   & ML   & Object tracking               & How to detect an object                                               \\ 
\hline
2015          & Giusti et al. \cite{giusti2015machine}               & ML   & Navigation                    & Image classification to determine direction                           \\ 
\hline
2018          & Kaufmann et al. \cite{kaufmann2018deep}              & ML   & Waypoint \& desired velocity  & How to detect an object                                               \\ 
\hline
2021          & Janousek et al. \cite{janousek2021deep}              & ML   & Landing \& flight planning    & How to recognize an object                                            \\ 
\hline
2023          & Vladov et al.~\cite{vladov2023modified}              & ML   & Stabilization                 & How to adjust controller parameters                                   \\ 
\hline \hline
2015          & Kim et al. \cite{kim2015deep}                        & DL   & Navigation                    & Image classification to assist in flights                             \\ 
\hline
2017          & Li et al. \cite{li2017deep}                          & DL   & Trajectory tracking           & Control signals                                                       \\ 
\hline
2017          & Smolyanskiy et al. \cite{smolyanskiy2017toward}      & DL   & Navigation                    & View orientation and lateral offset                                   \\ 
\hline
2018          & Jung et al. \cite{jung2018perception}                & DL   & Navigation                    & How to detect the center of a gate                                    \\ 
\hline
2018          & Loquercio et al. \cite{loquercio2018dronet}          & DL   & Navigation                    & How to adjust yaw angle, and probability of collision                 \\ 
\hline
2019          & Edhah et al. \cite{edhah2019deep}                    & DL   & Hovering                      & How to determine  propeller speed                                     \\ 
\hline
2019          & Mantegazza et al. \cite{mantegazza2019learning}      & DL   & Ground target tracking        & Image classification for control                                      \\ 
\hline
2023          & Cardenas et al.~\cite{cardenas2023intelligent}       & DL   & Position control              & How to determine the rotor speeds                                     \\ 
\hline \hline
2016          & Imanberdiyev et al. \cite{imanberdiyev2016autonomous}& RL   & Navigation                    & How to select the moving direction                                    \\ 
\hline
2017          & Polvara et al. \cite{polvara2017autonomous}          & RL   & Landing                       & How to detect a landmark and control vertical descent                 \\ 
\hline
2017          & Choi et al. \cite{choi2017inverse}                   & RL   & Trajectory tracking           & Control input                                                         \\ 
\hline
2017          & Kahn et al. \cite{kahn2017plato}                     & RL   & Avoiding failure              & Policy                                                                \\ 
\hline
2017          & Hwangbo et al. \cite{hwangbo2017control}             & RL   & Stabilization                 & How to determine the rotor thrusts                                    \\ 
\hline
2018          & Xu et al. \cite{xu2018monocular}                     & RL   & Landing                       & How to determine the velocities of the UAV                            \\ 
\hline
2018          & Lee at al. \cite{lee2018vision}                      & RL   & Landing                       & How to determine the roll and pitch angles                            \\ 
\hline
2018          & Vankadari et al. \cite{vankadari2018reinforcement}   & RL   & Landing                       & How to determine the velocities of the UAV on the x and y axis        \\ 
\hline
2018          & Kersandt et al. \cite{kersandt2018self}              & RL   & Navigation                    & How to select three actions: move forward, turn right, and turn left  \\ 
\hline
2018          & Pham et al. \cite{pham2018reinforcement}             & RL   & Navigation                    & How to select the  moving direction                                   \\ 
\hline
2019          & Rodriguez et al. \cite{rodriguez2019deep}            & RL   & Landing                       & How to determine velocities of the UAV on x and y axes                \\ 
\hline
2019          & Liu  et al. \cite{liu2019attitude}                   & RL   & Formation control             & Optimal control law                                                   \\ 
\hline
2019          & Lambert et al. \cite{lambert2019low}                 & RL   & Hovering                      & The mean and variance of the change in states                         \\ 
\hline
2019          & Manukyan et al. \cite{manukyan2019deep}              & RL   & Hovering                      & How to determine the rotor speeds                                     \\ 
\hline
2019          & Srivastava et al. \cite{srivastava2019least}         & RL   & Target tracking               & How to determine the velocities of the UAV on x, y, and z axes        \\ 
\hline
2019          & Wu et al. \cite{wu2019uav}                           & RL   & Trajectory planning           & How to select the moving direction                                    \\ 
\hline
2019          & Wang et al. \cite{wang2019autonomous}                & RL   & Navigation                    & How to determine the steering angle                                   \\ 
\hline
2019          & Zeng \& Xu \cite{zeng2019path}                       & RL   & Path Planning                 & How to select flight direction                                        \\ 
\hline
2020          & Yoo et al. \cite{yoo2020hybrid}                      & RL   & Trajectory tracking           & How to adjust PD and LQR controllers gains                            \\ 
\hline
2020          & Rubi et al. \cite{rubi2020deep}                      & RL   & Trajectory tracking           & How to determine the yaw angle                                        \\ 
\hline
2020          & Pi et al. \cite{pi2020low}                           & RL   & Trajectory tracking           & How to determine the rotor thrusts                                    \\ 
\hline
2020          & Zhao et al. \cite{zhao2020robust}                    & RL   & Formation control             & How to solve Bellman equation                                         \\ 
\hline
2020          & Guerra et al. \cite{guerra2020reinforcement}         & RL   & Trajectory optimization       & Control signal                                                        \\ 
\hline
2020          & Li et al. \cite{li2020uav}                           & RL   & Target tracking               & \begin{tabular}[c]{@{}c@{}}How to determine the angular velocity of yaw angle \\ \& Linear acceleration \end{tabular}          \\ \hline
2020          & Kulkarni et al. \cite{kulkarni2020uav}               & RL   & Navigation                    & How to select moving direction                                        \\ 
\hline
    2020          & Hu \& Wang \cite{hu2020proximal}                     & RL   & Speed optimization            & How to determine the rotor speeds                                 \\ 
    \hline
2021          & Kooi et al. \cite{kooi2021inclined}                  & RL   & Landing                       & How to determine the total thrust, and roll and pitch angles          \\ 
\hline
2021          & Rubi et al. \cite{rubi2021deep}                      & RL   & Trajectory tracking           & How to determine the yaw angle                                        \\ 
\hline
2021          & Bhan et al. \cite{bhan2021fault}                     & RL   & Avoiding failure              & How to adjust the gains of PD position controller                     \\ 
\hline
2021          & Li et al. \cite{li2021trajectory}                    & RL   & Trajectory planning           & How to obtain the parameter vector of the approximate value function  \\ 
\hline
2022          & Jiang et al. \cite{jiang2022deep}                    & RL   & Landing                       & How to determine the velocity of the UAV on x and y axes              \\ 
\hline
2022          & Abo et al. \cite{abo2022adaptive}                    & RL   & Landing                       & \begin{tabular}[c]{@{}c@{}}How to determine the roll, pitch, and yaw angles \\  and velocity of the UAV on z axis\end{tabular} \\ 
\hline
2022          & Panetsos et al. \cite{panetsos2022deep}              & RL   & Payload transportation        & How to obtain the reference Euler angles \& Velocity on z axis        \\ 
\hline
2022          & Ye et al. \cite{ye2022multi}                         & RL   & Navigation                    & How to select the moving direction and determine the velocity         \\ 
\hline
2022          & Wang \& Ye \cite{wang2022consciousness}              & RL   & Trajectory tracking           & How to determine the pitch and roll torques                           \\ 
\hline
2022          & Farsi \& Liu \cite{farsi2022structured}              & RL   & Hovering                      & Hot to determine the rotor speeds                                     \\ 
\hline
2023          & Xia et al. \cite{xia2023reinforcement}               & RL   & Landing                       & How to obtain the force and torque command                            \\ 
\hline
2023          & Ma et al. \cite{ma2023deep}                          & RL   & Trajectory tracking           & How to determine the rotor speeds                                     \\ 
\hline
2023          & Castro et al. \cite{castro2023adaptive}              & RL   & Path Planning                 & How to find optimized routes for navigation                           \\ 
\hline
2023          & Mitakidis et al. \cite{mitakidis2023deep}            & RL   & Target Tracking               & How to obtain the roll, pitch and yaw actions                         \\ 
\hline
2023          & Shurrab et al. \cite{shurrab2023reinforcement}       & RL   & Target localization           & How to determine the linear velocity and yaw angle                    \\ 
\hline
\end{tabular}}
\end{table*}

\subsection{Machine Learning} \label{subsec: machine learning}

Most offline ML techniques applied to and used for multirotor UAVs consider an onboard monocular camera. ML based approaches have been developed and adopted for navigation purposes, for stabilization, to track an object, to pass through waypoints with a desired velocity, and for landing purposes on stationary or dynamic targets. In addition, ML approaches are also used to tune and adjust controller parameters.  

In general, captured and acquired images are sent to pre-trained NNs, which first classify the obtained images into different classes, and then pass this information to the underlying multirotor controller as discussed in Bartak and Vykovsky~\cite{bartak2015any}, Janousek et al.~\cite{janousek2021deep}, Giusti et al.~\cite{giusti2015machine}, and Kaufmann et al.~\cite{kaufmann2018deep}. Specifics are offered next. 

Bartak and Vykovsky~\cite{bartak2015any} have combined computer vision, ML, and control theory techniques to develop a software tool for a UAV to track an object; the object is selected by a user who observes a series of video frames (images) and picks a specific object to be tracked by the multirotor. \textit{P-N learning}, where \textit{P} and \textit{N} represent positive and negative learning, respectively, is used by a Tracking-Learning-Detection (TLD) algorithm. The Lucas-Kanade tracker is implemented in the tracking phase, and a cascaded classification algorithm that includes a ML technique helps detect the object. A cascaded classification algorithm that consists of a patch variance classifier, an ensemble classifier, and a nearest neighbor classifier is also used. A simple RL algorithm decides the forward or backward speed of the multirotor by using a scaled size of the object. The yaw angle of the multirotor, used to follow the object, is provided to a Proportional–Integral–Derivative (PID) controller as input, to determine flight direction. 

Janousek et al.~\cite{janousek2021deep} have developed a method to accurately guide an autonomous UAV to land in a specific area that is labeled as the 'ground object'. The landing area includes a QR code, which, after it is identified and recognized it provides specific instructions/commands to the UAV for landing. A NN is used to identify the landing area using the onboard UAV camera. The recognition process is done on a ground control station, GCS, which is a part of the overall UAV-GCS ensemble. The GCS includes a communication channel for command transmission (to and from the UAV). False detection of the landing area may occur (i.e., due to sunlight) thus, success depends on how accurately the processed image determines the landing area, and not on the learning process itself. Regardless, when the UAV is within an 'acceptable flight altitude', a successful landing is accomplished.  

Giusti et al.~\cite{giusti2015machine} have used a quadrotor with an onboard monocular camera to determine the path configuration and direction of forest or mountain trails. A single image is collected from the onboard camera. A DNN is trained using supervised learning to classify obtained images. Two parameters are defined, $\Vec{v}$ for the direction of the camera's optical axis, and $\Vec{t}$ for the direction of the trail. Based on the calculated angle $\alpha$ between $\Vec{v}$ and $\Vec{t}$ and the angle $\beta$ that is $15^{\circ}$ around $\Vec{t}$, three actions are determined and classified as \textit{Turn Left (TL)}, \textit{Go Straight (GS)}, and \textit{Turn Right (TR)}, represented as \\
    {--} \textbf{TL} if $-90^{\circ} < \alpha < -\beta$ \\
    {--} \textbf{GS} if $-\beta \leq \alpha < +\beta$ \\
    {--} \textbf{TR} if $+\beta \leq \alpha < +90^{\circ}$ \\

These choices consider that the onboard camera centers and focuses on the motion direction. 

For performance evaluation and comparison, three alternatives are considered: learning using a Saliency-based model, learning following the method discussed in Santana et al.~\cite{santana2013tracking}, and by using two human observers who are asked to make one of the three previously mentioned decisions. The accuracy of the DNN is 85.2\%; the accuracy of the Saliency-based model is 52.3\%; the accuracy of the model in \cite{santana2013tracking} is 36.5\%. The accuracy of Human1 is slightly better than the accuracy of DNN, 86.5\%; Human2 has 82\% accuracy, which is lower than the accuracy of the DNN. This methodology has been tested experimentally and has produced successful results.  

Kaufmann et al.~\cite{kaufmann2018deep} have focused on the problem of autonomous, vision-based drone racing in dynamic environments with emphasis in path planning and control. A convolutional neural network (CNN) is used to detect the location of the waypoints from raw images and to decide about the speed to pass through the gates. The planner utilizes this information to design a short minimum jerk trajectory to reach the targeted waypoints. This technique is tested via simulations and in real environments. Comparisons with other navigation approaches and professional human drone pilots are made. It is shown that this method completes the track slower than human pilots do, but with a higher success rate. The success rate is also much higher compared to using visual-inertial odometry (VIO).

Vladov et al.~\cite{vladov2023modified} have studied the UAV stabilization problem. They adjust PID controller parameters using a recurrent multilayer perceptron (RMLP) method to stabilize the UAV attitude angles. The determined error is the input to a NN. Instead of using a constant training rate, an adaptive training rate is implemented to overcome slow convergence in the learning part and to avoid trapping in a local minimum during the learning phase. Results show that this method has a lower attitude error when compared to the RMLP method with a constant training rate and when using only an ANN.

\subsection{Deep Learning} \label{subsec: deep learning}

DL algorithms (discussed in 8 papers) that have been applied to multirotor UAVs focus on navigation and control, hovering, ground target tracking, and trajectory tracking. 

In Edhah et al.~\cite{edhah2019deep}, a DNN has been used to control UAV altitude and hover. The standard feedforward, greedy layer-wise, and Long Short-Term Memory (LSTM) methods are evaluated and compared. The controller outputs, position and speed errors, are collected every 1 ms, and then used by a supervised learning technique to train the DNN. After training, the trained DNN controller (and related parameters) is replaced by a Linear–Quadratic Regulator (LQR) controller. To overcome a slight offset in the output signal that results in a small error (in the system response), a proportional corrector is added in parallel to the DNN controller to recover the error in the DNN output signal. Best results are received when using the greedy layer-wise method.

Mantegazza et al.~\cite{mantegazza2019vision} have presented three different approaches and models to track a moving user. In the first model, the ResNet~\cite{he2016deep} architecture (a CNN) is utilized ~\cite{loquercio2018dronet}. Red-green-blue (RGB) images captured from 14 different people are used as input, providing x, y, and z positions as output. The second model follows the same structure, but velocities on the x, and y axes are provided as additional inputs. These additional inputs skip ResNet, but they are concatenated to the output of the NN. The outputs are control variables corresponding to four moving directions, up, down, left, and right. In the third model, a simple multilayer perceptron is utilized to map the quadrotor position on three axes and velocities on the x and y axes to control variables. The Mean Absolute Error (MAE) approach is deployed as a loss function during training of all three models. The first and second models use a simple baseline controller function. The last approach uses a combination of the first and third models. 

Li et al.~\cite{li2017deep} have studied trajectory tracking without any adaptation, but they have considered quadrotor stabilization and robustness in the presence of disturbances. A DNN is trained with labeled training examples using a standard feedforward method. The DNN uses the quadrotor desired trajectory and current states of position and translational velocities (on each axis), Euler angles, angular velocities, and acceleration on the z-axis. The quadrotor reference states are provided as the DNN output. The trained DNN is placed in front of a controller, and errors between current and desired states are used as inputs to a PID controller. Results show that the DNN with current state feedback is more efficient than the DNN without current state feedback. However, the DNN using future desired state feedback provides better performance.  

Kim and Chen~\cite{kim2015deep}, Jung et al.~\cite{jung2018perception}, Loquercio et al.~\cite{loquercio2018dronet}, and Smolyanskiy et al.~\cite{smolyanskiy2017toward} have centered around the quadrotor navigation task using DL techniques. 

Kim and Chen~\cite{kim2015deep}  have developed an autonomous indoor navigation system for a quadrotor to find a specific item using a single onboard camera. Six flight commands are used, Move Forward, Move Right, Move Left, Spin Right, Spin Left, and Stop. To establish a dataset, an expert pilot flies the quadrotor in seven different locations; images are collected from the UAV that are based on, corresponding to, (specific) flight commands. A CNN that is a modified CaffeNet model~\cite{CaffeMod98:online} is trained with the established dataset. This modification allows for faster training. During indoor flights, obtained images are classified by the trained NN, and based on image classification, specific control commands are issued to the quadrotor.

Jung et al.~\cite{jung2018perception} have developed a CNN to identify accurately the center of gates during indoor drone racing. ADRNet is built and trained using the Caffe library. To build the ADRNet, the AlexNet used in \cite{krizhevsky2017imagenet} and applied instead of the VGG-16 employed in \cite{simonyan2014very} are adopted. In \cite{liu2016ssd}, a convolutional layer is added among the fully connected layers of AlexNet instead of the fc6 and fc7 layers, while the fc8 layer is removed. Thus, a shorter inference time is required compared to the VGG-16-based Single-Shot-Detection (SSD) approach. ADRNet detects the center of the gate, and this information is forwarded to a Line-Of-Sight (LOS) guidance algorithm that issues specific flight control commands. Performance of three Single-Shot Detection (SSD) models, the VGG-16-based SSD, the AlexNet-based SSD, and the ADRNet are compared. ADRNet is the fastest model to detect the center of the gate. 

Different from the traditional map-localize-plan methods, Loquercio et al.~\cite{loquercio2018dronet} have applied a data-driven approach to overcome UAV challenges encountered when navigating in unstructured and dynamic environments. A CNN, called DroNet, which is used to navigate quadrotors through the streets of a city, is proposed. Collecting data within a city, in urban areas, to train UAVs, is dangerous for both, pedestrians and vehicles, even if an expert pilot flies the quadrotor. Therefore, publicly available datasets from Udacity's project are used to learn the steering angles. The dataset includes over 70,000 driving car images collected and classified through six experiments. Five experiments are for training and one for testing. A collision probability dataset is also developed for different areas of a city by placing a GoPro camera on the handlebars of a bicycle. UAV control is achieved via commands issued based on the output of DroNet. The collision probability is used to determine the quadrotor forward velocity. The desired yaw angle in the range $[-\frac{\pi}{2},\frac{\pi}{2}]$ is determined by using predicted scaled steering in the range $[-1,1]$. DroNet has worked successfully to avoid unexpected situations and obstacles, predicting the collision probability and the desired steering angle. The quadrotor learned to fly in several environments, and in indoor environments such as parking lots and corridors.

Smolyanskiy et al.~\cite{smolyanskiy2017toward} have focused on autonomously navigating a micro aerial vehicle (MAV) in unstructured outdoor environments. A DNN called TrailNet is developed and used to estimate the view orientation and lateral offset of the MAV with respect to the center of a trail. A DNN-based controller provides for a stable flight and avoids overconfident maneuvers by utilizing a loss function that includes both label smoothing and an entropy reward. The MAV includes two vision modules, a second DNN and a visual odometry component that is called direct sparse odometry (DSO). The second DNN helps detect the objects in the environment; the DSO estimates the depth to compute a pseudo-colored depth map. Their combination with TrailNet provides an autonomous flight controller functioning in unstructured environments.  ResNet-18, SqueezeNet (a miniature version of AlexNet), the DNN architecture in \cite{giusti2015machine}, and TrailNet are compared considering autonomous long-distance flight ability, prediction accuracy, computational efficiency, and power efficiency. Only Trailnet is 100\% autonomous; SqueezeNet and mini AlexNet are the closest to TrailNet with 98\% and 97\%, respectively. The least autonomous architecture is the DNN in \cite{giusti2015machine} with 80\%. Software modules run simultaneously in real-time, and the quadcopter successfully flies in unstructured environments. 

Cardenas et al.~\cite{cardenas2023intelligent} have developed a DNN-based flight position controller using a supervised DL technique. A dataset that includes position, velocity, acceleration, and motor output signals for different trajectories is created by using a PID flight controller. Five different NN architectures (from the literature) are utilized to learn the rotor speeds using a dataset, and their performance is compared. The five developed architectures are: $i)$ ANN, $ii)$ ANN Feedback, $iii)$ LSTM, $iv)$ LSTM Layers interleaved with convolutional 1D layers (LSTMCNN), and $v)$ Convolutional 1D Layers cascaded with LSTM layers (CLSTM). A comparative study shows that LSTMCNN gives the best performance as a DNN-based flight position controller. LSTMCNN performance is checked against the PID position controller, and it is shown that LSTMCNN has a wider operational range than the PID position controller. 

\subsection{Reinforcement Learning} \label{subsec: reinforcement learning}

\begin{table*}[]
    \centering
    \caption{Classification of Reinforcement Learning (Offline)}
    \label{Tab: Classification of RL}
    \resizebox{\textwidth}{!}{
    \begin{tabular}{|c|c|c|}
    \hline
    \textbf{Methods}         & \textbf{Algorithms}  & \textbf{Papers} \\
    \hline
                             & Q-learning   & Guerra et al.\cite{guerra2020reinforcement}, Pham et al.~\cite{pham2018reinforcement}, Kulkarni et al.~\cite{kulkarni2020uav}, Abo et al.\cite{abo2022adaptive}, Zeng \& Xu~\cite{zeng2019path} \\ 
                             & DQN          & Xu et al.~\cite{xu2018monocular}, Polvara et al.~\cite{polvara2017autonomous}, Castro et al.~\cite{castro2023adaptive}, Shurrab et al.~\cite{shurrab2023reinforcement}, Wu et al.~\cite{wu2019uav}, Kersandt et al.~\cite{kersandt2018self} \\
        Value Function-based & LSPI         & Vankadari et al.~\cite{vankadari2018reinforcement}, Lee et al.~\cite{lee2018vision}, Srivastava et al.~\cite{srivastava2019least} \\
                             & IRL          & Choi et al.~\cite{choi2017inverse} \\
                             & Others       & Imanberdiyev et al.~\cite{imanberdiyev2016autonomous}, Ye et al.~\cite{ye2022multi}, Farsi \& Liu~\cite{farsi2022structured}, Li et al.~\cite{li2021trajectory}, Xia et al.~\cite{xia2023reinforcement}  \\
    \hline                         
                            & PPO       & Kooi \& Babuška~\cite{kooi2021inclined}, Bhan et al~\cite{bhan2021fault}  \\ 
                            & TRPO      & Manukyan et al.~\cite{manukyan2019deep}  \\ 
        Policy Search-based & PILCO     & Yoo et al.~\cite{yoo2020hybrid}  \\
                            & PLATO     & Kahn et al.~\cite{kahn2017plato} \\
                            & Others    & Hu \& Wang~\cite{hu2020proximal}, Lambert et al.~\cite{lambert2019low} \\
    \hline                         
                            & DDPG  & Jiang \& Song~\cite{jiang2022deep}, Rodriguez et al.~\cite{rodriguez2019deep}, Rubi et al.~\cite{rubi2020deep}, Rubi et al.~\cite{rubi2021deep}, Ma et al.~\cite{ma2023deep}, Mitakidis et al.~\cite{mitakidis2023deep} \\ 
                            & TD3   & Jiang \& Song~\cite{jiang2022deep}, Kooi \& Babuška~\cite{kooi2021inclined}, Li et al.~\cite{li2020uav}, Panetsos et al.~\cite{panetsos2022deep} \\ 
                            & SAC   & Jiang \& Song~\cite{jiang2022deep}, Kooi \& Babuška~\cite{kooi2021inclined},  \\
        Actor-Critic        & Fast-RDPG & Wang et al.~\cite{wang2019autonomous} \\
                            & DeFRA & Li et al.~\cite{li2021lstm} \\
                            & CdRL & Wang \& Ye~\cite{wang2022consciousness} \\
                            & Others & Pi et al.~\cite{pi2020low}, Hwangbo et al.~\cite{hwangbo2017control} \\
    \hline
    \end{tabular}}
\end{table*}

Offline RL has been extensively applied to multirotor UAVs. The literature review reveals 42 papers that focus on 13 different tasks, which include trajectory tracking, landing, navigation, formation control, flight control, and hovering.  

\begin{figure}
\begin{center}
\includegraphics[width=10.5 cm]{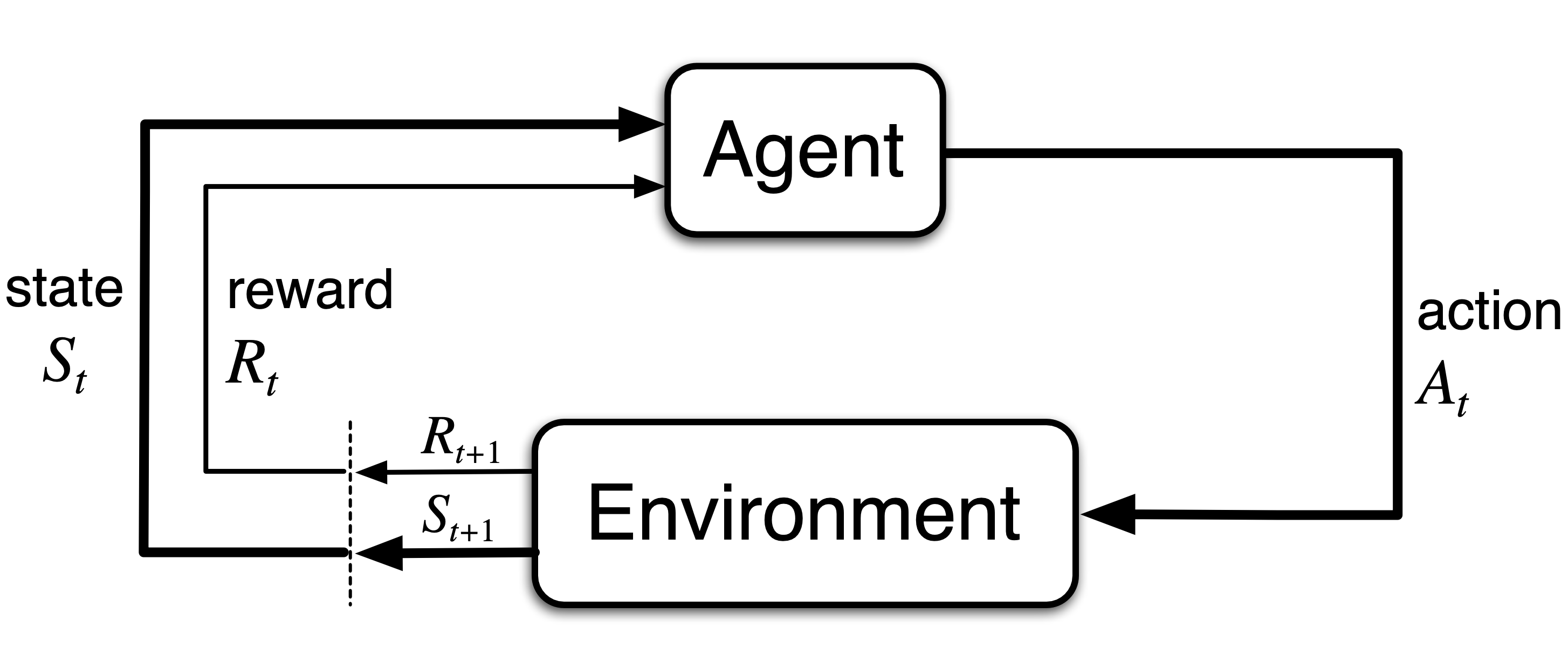}
\caption{Block diagram and interaction between agent and environment \cite{sutton2018reinforcement}.}
\label{RL_interaction}
\end{center}
\end{figure}

In general, the RL framework uses an agent that is trained through trial-and-error to decide on an action that maximizes a long-term benefit. RL is described by a Markov Decision Process (MDP). The agent-environment interaction in an MDP is illustrated in Fig.~\ref{RL_interaction}, where, Agent, Environment, and action represent in engineering terms the controller, the controlled system, and the control signal, respectively \cite{sutton2018reinforcement}.

RL algorithms are classified according to whether they are model-based or model-free, on-policy or off-policy, value function-based or policy search-based, or according to whether they are derived for planning or learning purposes. In what follows, the classification is in terms of value function-based, policy search-based, or actor-critic. Table~\ref{Tab: Classification of RL} offers a summary of the different variations of RL algorithms.

\subsubsection{Value Function-Based Algorithms} \label{subsubsec: value function-based}

Value function-based methods use state-value and action-state functions that are presented in (\ref{eq: state-value function}) and (\ref{eq: action-state function}), respectively, shown next  

\begin{equation} \label{eq: state-value function}
    v_\pi(s) = \mathbb{E}_\pi \left[\sum_{k=0}^{\infty}\gamma^{k}R_{t+k+1}\ \middle|\ S_{t} = s \right] for \ all \ s \in S 
\end{equation}    

\begin{equation} \label{eq: action-state function}
    q_\pi(s,a) = \mathbb{E}_\pi \left[\sum_{k=0}^{\infty}\gamma^{k}R_{t+k+1}\ \middle|\ S_{t} = s, A_{t} = a \right]
\end{equation}
where $v_\pi(s)$ denotes the \textit{value function} for policy $\pi$ at state \textit{s} while $q_\pi(s,a)$ represents the \textit{action-value function} for policy $\pi$ at state \textit{s} and action \textit{a}. $\mathbb{E}_{\pi}[\cdot]$ denotes the expected value under policy $\pi$. $\sum_{k=0}^{\infty}\gamma^{k}R_{t+k+1}$ is the sum of discounted future rewards starting from time \textit{t} in state \textit{s} and represents the expected \textit{discounted return}, and $\gamma$ is the \textit{discount rate}, $0 \leq \gamma \leq 1$, here.  $S_t$ and $A_t$ represent the state and action at time \textit{t}, respectively \cite{sutton2018reinforcement}\cite{bilgin2020mastering}\cite{lapan2018deep}.

The discount rate $\gamma$ plays a critical role in calculating the present value of future rewards. If $\gamma < 1$, the infinite sum has a finite value when the reward sequence, $R_{k}$, is bounded. If $\gamma = 0$, the agent cannot calculate future rewards; it can only calculate the immediate reward ($\sum_{k=0}^{\infty}\gamma^{k}R_{t+k+1} = 0^0R_{t+1} + 0^1R_{t+2} + \cdots = R_{t+1}$). So the agent learns how to choose the action $A_t$ to maximize $R_{t+1}$. If $\gamma$ approaches 1, the future rewards are highlighted in the expected discount return, that is, the agent behaves more farsighted. For example, in \cite{sutton2018reinforcement} the discount factor is chosen close to 1. Li et al.~\cite{li2021trajectory}, Hu and Wang~\cite{hu2020proximal} have chosen the value of 0.9, and Castro et al.~\cite{castro2023adaptive} and Panetsos et al~\cite{panetsos2022deep} have chosen the discount factor value to be 0.99.

Value function-based algorithms consist of variance algorithms, such as Dynamic Programming (DP), Monte Carlo (MC), and Temporal-Difference (TD). Two most popular DP methods, policy iteration and value iteration, benefit from policy evaluation and policy improvement. The MC method is also based on policy evaluation and policy improvement, but unlike the DP method, an alternative policy evaluation process is utilized. The policy evaluation in the DP method employs a bootstrapping technique, while sampling and average return techniques are applied in the MC method. The TD method is created by combining the DP and MC methods applying sampling and bootstrapping \cite{sutton2018reinforcement}\cite{bilgin2020mastering}\cite{lapan2018deep}.

TD has been extensively applied in control algorithms for multirotors executing diverse tasks like landing, navigation, obstacle avoidance, path planning, and trajectory optimization. Q-learning and Deep Q-Networks (DQN) stand out as commonly employed RL algorithms within the TD framework, particularly in value function-based algorithms, see Guerra et al.~\cite{guerra2020reinforcement}, Pham et al.~\cite{pham2018reinforcement}, and Polvara et al.~\cite{polvara2017autonomous}.

Xu et al.~\cite{xu2018monocular} have applied an end-to-end control scheme that includes a DNN and a double DQN algorithm for quadrotor landing on a stable platform. The output of the underlying Deep Reinforcement Learning (DRL) model is the quadrotor speed in x and y - the velocity in the z direction is not controlled, it is considered fixed; this makes the problem easier. After testing, the improved DQN method produces good results on autonomous landing.  

Imanberdiyev et al.~\cite{imanberdiyev2016autonomous} have used a model-based RL algorithm on a quadrotor to create an efficient path to reach a destination by considering its battery life. The agent uses three states, position in x and y, and the battery level. The agent follows one of eight possible actions, moving on the x-y plane, and learning the moving direction. The direction action is converted to trajectory commands that are executed using position control. In model-based RL methods, there is a limited number of actions the agent learns to create a sufficiently accurate environment model, as opposed to model-free RL methods. Model-based RL algorithms are not suitable for real-time systems since the planning and model learning aspects are computationally expensive. However, in \cite{imanberdiyev2016autonomous} a parallel architecture is used, called TEXPLORE. Thus, it is possible to take action fast enough based on the current policy - there is no need to wait for the planning and model update. Simulation results illustrate that the approach has the ability to learn and to perform well after a few iterations and to also perform actions in real-time. Performance is compared with Q-learning algorithms. In 150 episodes, while there is no significant change in the average reward in Q-learning, the average reward of TEXPLORE dramatically increases after the $25^{th}$ episode. TEXPLORE obtains significantly more rewards than Q-learning in each episode.

In \cite{zeng2019path}, the path design problem for a cellular-connected UAV has been handled to reduce mission completion time. A new RL-based UAV path planning algorithm is derived and TD is applied to learn directly the state-value function. A linear function is added to the algorithm with tile coding. Function approximation has two advantages over a table-based RL. It learns the parameter vector that has a lower dimension than the state vector, instead of storing and updating the value function for all states. It also allows for generalization. Tile coding is used to build the feature vector. The parameter vector may be updated to minimize its mean squared error based on a stochastic semi-gradient method with a linear approximation for each state-reward-nextState transition observed by the agent. It is shown that TD with a tile coding algorithm overcomes problems with cellular networks in complex urban environments of size 2 km $\times$ 2 km with high-rise buildings. Also, accumulated rewards from the TD and the TD with tile coding learning algorithms are almost identical, but tile coding provides faster convergence. When tested, the UAV reaches the desired location without running into the coverage holes of cellular networks.

The approach discussed in \cite{polvara2017autonomous} has used two DQNs. One is utilized for landmark detection, the other is used to control the UAV vertical descent. A hierarchy representing sub-policies is applied to the DQNs to reach decisions during the different navigation phases. The DQNs can autonomously decide about the next state. However, the hierarchy decreases the sophistication of the task decision. Algorithm performance is compared with an augmented reality (AR) tracker algorithm and with human pilots. The proposed algorithm is faster than a human pilot when landing on a marked pad but also more robust than the AR tracker in finding the marker. Complete details of \cite{polvara2017autonomous} may be found in \cite{polvara2020sim}.

Ye et al.~\cite{ye2022multi} have developed a DRL-based control algorithm to navigate a UAV swarm around an unexplored environment under partial observations. This may be accomplished by using GAT-based FANET (GAT-FANET), which is a combination of the flying Ad-hoc network (FANET) and the graph attention network (GAT). Partial observations lead to loss of information. Thus, a network architecture named Deep Recurrent Graph Network (DRGN) is developed and combined with GAT-FANET to collect environment spatial information, and use previous information from memory via a gated recurrent unit (GRU). A maximum-entropy RL algorithm, called soft deep recurrent graph network (SDRGN), is developed, which is a multi-agent deep RL algorithm. It learns a DRGN-based stochastic policy with a soft Bellman function. The performance of the DRGN (a deterministic model) and the SDRGN are compared with DQN, multi-actor attention critic (MAAC), CommNet, and graph convolutional RL (DGN). In a partially observable environment, the stochastic policy approach is more robust than the deterministic policy one. Also, GAT-FENAT provides an advantage because of its memory unit. When the number of UAVs increases, more information is required from the GAT-FANET, and this reduces dependency on the memory unit. Results \cite{ye2022multi} show that policies based on GAT-FANET provide better performance on coverage than other policies. It is observed that graph-based communication improves performance in cooperative exploration and path planning, too. The SDRGN algorithm has lower energy consumption than DRGN, but DQN has the lowest energy consumption when compared with DRL methods. SDRGN and DRGN performance increases linearly with the number of UAVs. SDRGN shows better performance than DRL methods - this verifies that it has better transferability. Consequently, overall, SDRGN has better performance, scalability, transferability, robustness, and interpretability than other DRL methods. 

Abo et al.~\cite{abo2022adaptive} have solved the problem of UAV landing on a dynamic platform taking advantage of Q-learning. Two types of adaptive multi-level quantization (AMLQ) are used; AMLQ 4A with 4 actions and AMLQ 5A with 5 actions, and then compared with a PID controller. The PID position magnitude errors in x and y are higher than the corresponding AMLQ errors, while the oscillation in the AMLQ models is higher than in the PID controller. The developed AMLQ reduces the error on the targeted landing platform. This solution provides faster training and allows for knowledge representation without the need of a DNN.

Path planning is effectively used in several areas that include precision agriculture. Castro et al.~\cite{castro2023adaptive} have worked on adaptive path planning using DRL to inspect insect traps on olive trees. The proposed path planning algorithm includes two parts, the rapidly-exploring random tree (RRT) algorithm and a DQN algorithm. The former searches for path options, while the latter performs optimized route planning integrating environment changes in real-time; however, the training process of DQN is completed offline. Simulation runs are done in an area of 300 $m^2$ with 10 dynamic objects; the UAV is provided with a safe route determined by the proposed approach, and it arrives at the insect traps to take their picture. 

Shurrab et al.~\cite{shurrab2023reinforcement} have studied the target localization problem and have proposed a Q-learning-based data-driven method in which a DQN algorithm helps overcome dimensionality challenges. Data measurements, from previous and the current step, the previous action, and the direction of the nearest boundary of the UAV compose the state space. The action space includes the UAV linear velocity and the yaw angle that determines flight direction. This approach is compared with the traditional uniform search method and the gradient descent-based ML technique; it returns better results in terms of localization and traveled distance.

Guera et al.~\cite{guerra2020reinforcement} have emphasized detection and mapping for trajectory optimization. For detection, the aim is to minimize 'wrong detection', and for mapping, the aim is to minimize the uncertainty related to estimating the unknown environment map. The proposed MDP-based RL algorithm, inspired by Q-learning, consists of state and control estimations. The states are the UAV position depending on actions, a binary parameter that shows the presence or absence of a signal source in the environment, and the states of each cell. The action space includes the control signal to move the UAV from one cell to another in the grid (environment) map. Numerical results show that this technique provides a high probability of target detection and improves the capabilities of map exploration. 

Pham et al.~\cite{pham2018reinforcement} have handled the UAV navigation problem using Q-learning. The navigation problem is formulated using a discretized state space within a bounded environment. The algorithm learns the action, which is the UAV moving direction in the described environment. The state space includes the distance between the UAV and the target position, and the distance to the nearest obstacle in North, South, West, or East directions. UAV navigation following the shortest path is demonstrated. 

Kulkarni et al.~\cite{kulkarni2020uav} have also used Q-learning for navigation purposes. The objective is to determine the location of a victim by using a RF signal emitted from smart devices. The transmitted signal reaches the agent and according to the received signal strength (RSS), the agent learns to choose one of eight directions separated by 45 degrees on the x-y plane. For mapping, a grid system is utilized, and each state label is correlated to a particular RSS value (two adjacent grids in the map have different RSS values). Each location on the map has a unique state. The $\epsilon$-greedy approach provides an action to the UAV, and each episode or iteration is completed when the RSS value of the grid is determined to be greater than -21dBm  - this value means that the distance from the victim is less than 2 meters. The proposed approach is tested for different starting positions on different floor plans, demonstrating that the UAV successfully reaches the victim's position.

Choi et al.~\cite{choi2017inverse} have trained a multirotor UAV by mimicking the control performance of an expert pilot. A pilot collects data from several actual flights. Then, a hidden Markov model (HMM) and dynamic time warping (DTW) are utilized to create the trajectory. Inverse RL is used to learn the hidden reward function and use it to design a controller for trajectory following. Simulations and experiments show successful results.

Wu et al.~\cite{wu2019uav} have worked on the general task of 'object finding', for example, in rescue missions. A DQN algorithm is used for trajectory planning. The elimination of the loop storm effect that reflects the current sequence in MDP is repeated and it does not cause a punishment in continued actions. The Odor Storm that is caused by not reaching the highest reward value when the agent gets closer to the target increases the convergence speed of the training process. It is shown that the break loop storm technique and the odor effect reduce the training process time.

In Kersandt et al.~\cite{kersandt2018self} a DNN has been trained with a DRL algorithm to control a fully autonomous quadcopter that is equipped with a stereo-vision camera to avoid obstacles. Three different DRL algorithms, DQN, Double DQN (DDQN), and Dueling DDQN are  applied to the system. The average performance of each algorithm with respect to rewards is 33, 120, and 116, respectively; they all are below human performance. The results of applying DDQN and Dueling techniques show that the quadrotor reaches the target with 80\% success. 

Liu et al.~\cite{liu2019attitude} and Zhao et al.~\cite{zhao2020robust} have used RL for formation control. In  \cite{liu2019attitude}, a value function-based RL control algorithm is applied to leader-follower quadrotors to tackle the attitude synchronization problem. The output of each quadrotor is synchronized with the output of the leader quadrotor by the designed control system. In \cite{zhao2020robust}, the aim is to solve the model-free robust optimal formation control problem by utilizing off-policy value function-based algorithms. The algorithms are trained for robust optimal position and attitude controllers by using the input and output data of the quadrotors. Theoretical analysis and simulation results match, and the robust formation control method works effectively. 

Vankadari et al.~\cite{vankadari2018reinforcement} and Lee et al.~\cite{lee2018vision} have worked on the landing task using a Least Square Policy Iteration (LSPI) algorithm that is considered a form of approximate dynamic programming (ADP).  Srivastava et al.~\cite{srivastava2019least} and Li et al.~\cite{li2021trajectory} have applied a LSPI algorithm in multirotor UAVs for target tracking and trajectory planning, respectively. ADP is used to solve problems with large state or action spaces. ADP approximates the value function or policy with function approximation techniques, since storing values for every state-action pair is not practical. 

In \cite{vankadari2018reinforcement}, a LSPI algorithm has been used to study the landing problem. A RL algorithm estimates quadrotor control velocities using instantaneous position and velocity errors. The optimal value function of any policy is determined by solving the Bellman equation as applied to a linear system. In the RL algorithm, the LSPI method forecasts the value function parameterizing it into basis functions instead of calculating an optimal value function. The RL algorithm converges quickly and learns how to minimize the tracking error for a given set point. Different waypoints are used to train the algorithm for landing. The method can also be used in noisy environments, effectively. Simulations and real environment results demonstrate applicability of the approach. 

Research in \cite{farsi2022structured} has provided a low-level control approach for a quadrotor by implementing a structured online learning-based algorithm (SOL) \cite{farsi2020structured} to fly and keep the hovering position at a desired altitude. The learning procedure consists of two stages; the quadrotor is first flown with almost equal pulse-width modulation (PWM) values for each rotor; these values are collected to create an initial model. Then, learning in a closed-loop form is applied using the initial model. Before applying closed-loop learning, three pre-run flights are completed. 634 samples are collected in 68 seconds of flying. The state samples are determined at each time step in the control loop, and then the system model is updated using an RLS algorithm. After determining the updated model, a value function (needed to find the control value for the next step) is updated. The quadrotor is autonomously controlled. This online learning control approach successfully reaches the desired position and keeps the quadrotor hovering.  

In \cite{lee2018vision} a trained NN has been adopted for guidance in a simulation environment. A quadrotor with a PID controller has been used, which has onboard a ground-looking camera. The camera provides pixel deviation of the targeted landing platform from an image frame, and a laser rangefinder that procures the altitude information. During training, the NN is trained to learn how to control the UAV attitude. In simulation studies, the UAV reaches the proposed landing location. In experiments, the AI pilot is turned off below the altitude of 1.5 m, but the AI pilot can land at the targeted location using a vision sensor. Trajectories are not smooth because the landing location in the image is not accurately determined due to oscillations, because of image processing errors in the actor NN, because  signal transmission creates a total delay of 200 ms, and because of disturbances in real-world environments. 

Three target tracking approaches that deserve attention are the Image-based (IBVS), Position-based (PBVS), and Direct Visual Servoing approaches. In Kanellakis and Nikolakopoulos~\cite{kanellakis2017survey}, IBVS has been found to be the more effective approach for target tracking since it tackles directly the control problem in the image space; it also has better robustness when it comes to camera calibration and to depth estimation errors. Srivastava et al.~\cite{srivastava2019least} track a maneuvering target using only vision-based feedback, IBVS. However, tracking is difficult when using only monocular vision without depth measurements. This deficiency is eliminated by a RL technique where optimal control policies are learned by LSPI to track the target quadrotor. Two different basis functions (with and without velocity basis), and four types of reward functions (only exponential reward, quadratic reward function without velocity control, quadratic reward function with velocity control, asymmetric reward function) are described in \cite{srivastava2019least}. The basis function with velocity basis shows better performance than the basis function without velocity basis.

In \cite{li2021trajectory}, the objective has been to solve the problem of cable-suspended load transportation utilizing three quadrotors. The trajectory planning method is based on a value-function approximation algorithm with the aim to reach the final position as fast as possible keeping the load stable. This method includes two processes, trajectory planning and tracking. The trajectory planning process consists of parameter learning and trajectory generation. Training and learning help determine the parameter vector of the approximate value function (parameters learning part). In the trajectory generation phase, the value function is approximated by using the learned parameters in the former stage, and the flight trajectory is determined via a greedy strategy. The results effectiveness of the load trajectory and the physical effect on the quadrotor flight are checked based on the trajectory tracking process. The quadrotors are independent; in the trajectory tracking phase, positions and attitudes are controlled with a hierarchical control scheme using PID controllers (transmitting the position of the load to the controller of the quadrotor). Results show that the actual value function is successfully estimated. Also, the value function confirms that the proposed algorithm works effectively.

Xia et al.~\cite{xia2023reinforcement} have used a RL control method for autonomous landing on a dynamic target. Unlike other studies, position and orientation constraints for safe and accurate landing are described. Adaptive learning and a cascaded dynamic estimator are utilized to create a robust RL control algorithm. In the adaptive learning part, the critic network weight is formulated and calculated in an adaptive way. Also, the stability of the closed-loop system is analyzed.

\subsubsection{Policy Search-Based Algorithms} \label{subsubsec: policy search-based}

Value function-based methods calculate the value of an agent's every possible action to choose the one based on the best value. The probability distribution over all available actions plays a key role in policy-based methods, and for the agent to decide the action each time step. A comparison of value function-based and policy search-based algorithms is provided in Table~\ref{Tab: Comparison of value and policy}.

\begin{table}[]
    \centering
    \caption{Comparison of value function-based and policy search-based methods}
    \label{Tab: Comparison of value and policy}
    \begin{tabular}{c|c}
        \textbf{Value Function-Based} & \textbf{Policy Search-Based} \\
        \hline
        Indirect policy optimization & Direct policy optimization \\
        Generally off-policy & On-policy \\
        Simpler algorithm & Complex algorithm \\
        Computationally expensive & Computationally inexpensive \\
        More iteration to converge & Less iteration to converge
    \end{tabular}
\end{table}

Kooi and Babuška~\cite{kooi2021inclined} have developed an approach using deep RL to land a quadrotor on an inclined surface, autonomously. Proximal Policy Optimization (PPO), Twin-Delay Deep Deterministic Gradient (TD3), and Soft Action-Critic (SAC) algorithms are applied to solve this problem. The TD3 and SAC algorithms trained successfully the set-point tracking policy network, but the PPO algorithm was trained in a shorter time and provided a better performance on the final policy. Trained policies may be implemented in real-time. 

Hu and Wang~\cite{hu2020proximal} have utilized an advanced PPO RL algorithm to find the optimal stochastic control strategy for a quadrotor speed. During training, an actor and a critic NN are used. They have the same nine-dimensional state vector (Euler angles, Euler angle derivatives, errors between expected and current velocities after integration on the x, y, and z axes). An integral compensator is applied to both NNs to improve speed-tracking accuracy and robustness. The learning approach includes online and offline components. In the offline learning phase, a flight control strategy is learned using a simplified quadrotor model, which is continuously optimized in the online learning phase. In offline learning, the critic NN evaluates the current action to determine an advantage value choosing a higher learning rate to improve evaluation. In online learning, the action NN is composed of four policy trainable sub-networks. The state vector is used as input to the four sub-networks; their outputs are the mean and variance of the corresponding four Gaussian distributions, each normalized to [0, 1]. Parameters of the four policy sub-networks are also used in the old policy networks that are untrainable. The old policy sub-network parameters are fixed. The four policy sub-networks in the action NN are trained to produce new actions in the next batch. When applying new actions to the quadrotor, new states are recorded in a buffer. After the integration and compensation process, a batch of the state vector is used as input to the critic NN. The batch of the advantage values is the output of the critic NN; it is used to evaluate the quality of the actions taken to determine these states. The parameters of the critic NN are updated by minimizing the advantage value per batch. The policy network is updated per batch using the action vectors taken from the old policy network, the state vector from the buffer, and the advantage value from the critic NN. 

In \cite{hu2020proximal}, the PPO and the PPO-IC algorithms have been compared with the offline PPO one and a well-tuned PID controller. The average linear velocity steady-state error of the PPO-IC approaches zero faster, and it is smaller than the PPO. The average accumulated reward of the PPO-IC reaches a higher value. The PPO-IC converges closer to the targeted velocity in the x, y, and z axes than the PPO. PPO-IC velocity errors in the x, y, and z axes are much smaller compared to the PPO errors. The Euler angle errors are also smaller in the PPO-IC algorithm. In the offline learning phase, the nominal quadrotor weight is increased by 10\% in each step until it reaches 150\% of the nominal weight. The performance of the well-tuned PID controller and the proposed method are compared. When the quadrotor weight increases, the velocity error along the z-axis increases, too, but the PPO-IC algorithm demonstrates stable behavior without fluctuations in speed tracking. Moreover, 12 experiments are conducted when the nominal 0.2 m radius of the quadrotor is increased from 50\% to 550\%, that is, from 0.1 m to 1.1 m. PID and PPO-IC performance are similar when the radius is between 0.2 - 0.4 m. For higher values, PID performance decreases even when convergence to the desired value is observed. However, the PID controller cannot control the quadrotor - it becomes unstable when the radius increases to more than 1 m. On the contrary, changes in the radius value slightly affect the performance of the PPO-IC algorithm.

Kahn et al.~\cite{kahn2017plato} and Bhan et al.~\cite{bhan2021fault} have worked on failure avoidance and on compensating for occurred failures during flights. In \cite{kahn2017plato}, a Policy Learning using Adaptive Trajectory Optimization (PLATO) algorithm, a continuous, reset-free RL algorithm, is developed. In PLATO, complex control policies are trained with supervised learning using model predictive control (MPC) to observe the environment. Partially trained and unsafe policies are not utilized in the action decision. During training, taking advantage of the MPC robustness, catastrophic failures are minimized since it is not necessary to run the learned NN policy during training time. It is shown that good long-horizon performance of the resulting policy is achieved by the adaptive MPC. In \cite{bhan2021fault}, accommodation and recovery from fault problems occurring in an octacopter is achieved using a combination of parameter estimation, RL, and model-based control. Fault-related parameters are estimated using an Unscented Kalman Filter (UKF) or a Particle Filter (PF). These fault-related parameters are given as inputs to a DRL, and the action NN in the DRL provides the new set of control parameters. In this way, the PID controller is updated when the control performance is affected by the parameter(s) correlated with faults.

In \cite{manukyan2019deep}, a DRL technique has been applied to a hexacopter to learn stable hovering in a state action environment. The DRL used for training is a model-free, on-policy, actor-critic-based algorithm called Trust Region Policy Optimization (TRPO). Two NNs are used as nonlinear function approximators. Experiments show that such a learning approach achieves successful results and facilitates controller design.

Yoo et al.~\cite{yoo2020hybrid} have combined RL and deterministic controllers to control a quadrotor. Five different methods, the original probabilistic inference for learning control (PILCO), PD-RL with high gain, PD-RL with low gain, LQR-RL, and LQR-RL with model uncertainty are compared via simulations when the quadrotor tracks a circular reference trajectory. The high-gain PD-RL approaches fast the reference trajectory. The low-gain PD-RL behaves less aggressively and reference trajectory tracking is delayed. The convergence rate of the PD-RL and LQR-RL methods is better. Performance is also better when compared to the original PILCO. The main advantages of combining a deterministic controller with PILCO are simplicity and rapid learning convergence. 

In \cite{lambert2019low}, errors on the pith and roll angles are minimized to provide stability at hovering. A user-designed objective function uses simulated trajectories to choose the best action. The objective function also minimizes the cost of each state. The performance of this controller is worse than a typical quadrotor controller performance. However, the proposed controller achieves hovering for up to 6 seconds after training using 3 minutes of data.

\subsubsection{Actor-Critic Algorithms} \label{subsubsec: actor-critic}

Actor-critic algorithms consist of both value function-based and policy search-based methods. The actor refers to the policy search-based method and chooses the actions in the environment; the critic refers to the value function-based method and evaluates the actor using the value function.

In \cite{jiang2022deep}, three different RL algorithms, DDPG, TD3, and SAC, have been applied to study multirotor landing. Using the DDPG method does not result in successful landings. The TD3 and SAC methods successfully complete the landing task. However, TD3 requires a longer training period and landing is not as smooth, most likely because of noise presence in the algorithm.   

Rodriguez et al.~\cite{rodriguez2019deep} have studied landing on a dynamic/moving platform using DDPG. Slow and fast scenarios have been tried in 150 test episodes. During the slow scenario, the moving platform (the moving platform trajectory is periodic) velocity is 0.4 m/s, and during the fast scenario is set at 1.2 m/s. The success rate is 90\% and 78\%, respectively. Using a constant velocity on the z-axis results in landing failure on the moving platform. This problem may be overcome by using the velocity on the z-axis as a state, but this makes the training process more complicated and learning the landing process becomes more challenging.

Rubi et al.~\cite{rubi2020deep} have solved the quadrotor path following problem using a deep deterministic policy gradient (DDPG) reinforcement learning algorithm. A lemniscate and one lap of a spiral path are used to compare agents with different state configurations in DDPG and in an adaptive Nonlinear Guidance Law (NLGL) algorithm. The agent has only two states, distance error and angle error. According to the results, the adaptive NLGL has a lower distance error than the 2-state agent, but its distance error is significantly greater than the agent with the future states on the lemniscate path. 

Rubi et al.~\cite{rubi2021deep} have also used three different approaches to solve the path following problem using DDPG. The first agent utilizes only instantaneous information, the second uses a structure (the agent expects the curve), and the third agent computes the optimal speed according to the shape of the path. The lemniscate and spiral paths are used to test the three agents. The lemniscate path is used in the training and test phases. The agents are evaluated in tests but with the assumption that the third agent is also limited by a maximum velocity of 1 m/s. For the lemniscate path, the agents are first tested with ground truth measurements. The second agent shows the best performance with respect to cross-track error. When the agents are tested with the sensor model, the third agent shows slightly better performance in terms of cross-track errors. Then, all agents are tested in the spiral path. When the performance of the agents is compared in simulations with ground truth measurements and with sensor models, the third agent (with a maximum velocity of 1 m/s) shows the best performance in terms of position error. In all tests, the third agent (without a maximum velocity limitation) completes the tracks faster.   

Wang et al.~\cite{wang2019autonomous} have handled the UAV navigation problem in a large-scale environment using a DRL. Two policy gradient theorems within the actor-critic framework are derived to solve the problem that has been formulated as a partially observable Markov decision process (POMDP). As opposed to conventional navigation methods, raw sensor measurements are utilized in the DRL; control signals are the output of the navigation algorithm. Stochastic and deterministic policy gradients for POMDP are applied to the RL algorithm. The stochastic policy requires samples from both the state and action space. The deterministic policy requires only samples from the state space. Therefore, the RL algorithm with a deterministic policy is faster (and preferred) - it is called a fast recurrent deterministic policy gradient algorithm (Fast-RDPG). For comparisons, four different large-scale complex environments are built with random-height buildings to test the DDPG, RDPG, and Fast-RDPG. The success rate of the Fast-RDPG is significantly higher in all environments. Fast-RDPG has the lowest crash rate in one environment. DDPG provides the best performance with respect to the average crash rate in all environments. Fast-RDPG has a much lower crash rate than RDPG. However, Fast-RDPG provides a much lower stray rate than other algorithms in all environments. 

Li et al.~\cite{li2021lstm} have developed a new DRL-based flight resource allocation framework (DeFRA) for a typical UAV-assisted wireless sensor network used for smart farming (crop growth condition). The DeFRA reduces the overall data packet loss in a continuous action space. A DDPG is used in DeFRA, and DeFRA learns to determine the instantaneous heading and speed of the quadrotor and to choose the ground device to collect data from the field. The time-varying airborne channels and energy arrivals at ground devices cause variations in the network dynamics. The network dynamics are estimated by a newly developed state characterization layer based on LSTM in DeFRA. An MDP handles simultaneously the control of the quadrotor’s maneuver and the communication schedule according to decision parameters (time-varying energy harvesting, packet arrival, and channel fading). The state space comprises the battery level, the data buffer length of all ground devices, the battery level and location of the UAV, the channel gain between the UAV and the ground devices, and the time-span parameter of the ground device. The UAV current battery level depends on the battery level of the UAV in the previous time step, harvested energy, and energy consumption. The quadrotor is required to keep its battery level at an equal or higher than the battery level threshold. Performance is compared with two DRL-based policies, DDPG-based Movement Control (DDPG-MC) and DQNs-based Flight Resource Allocation Policy (DQN-FRA) and with two non-learning heuristics, Channel-Aware Waypoint Selection (CAWS) and Planned Trajectory Random Scheduling (PTRS). The DeFRA framework provides lower packet loss than other methods. The relation between the packet loss rate and the number of ground devices is investigated according to all methods. The DRL-based methods outperform CAWS and PTRS. For up to 150 ground devices, DeFRA and DDPG-MC show similar performance and better than other methods, but after increasing the number of ground devices to 300, DeFRA provides better performance than DDPG-MC.

Pi et al.~\cite{pi2020low} have created a low-level quadrotor control algorithm to hover at a fixed point and to track a circular trajectory using a model-free RL algorithm. The combination of the on-policy and off-policy methods is used to train an agent. The standard policy gradient method determines the update direction within the parameter space, while the TRPO and PPO algorithms are designed to identify an appropriate update size. However, for updating the policy, the proposed model establishes new updating criteria that extend beyond the parameter space concentrating on local improvement. The NN output provides the thrust of each rotor. The simulator is created in Python using the dynamic model of Stevens et al.~\cite{stevens2015aircraft}. The effects of the rotation matrix and quaternion are investigated in the learning process. The model with the quaternion may converge slower in the training process than the model with the rotation matrix. However, both models show similar performance when tested.

Ma et al.~\cite{ma2023deep} have developed a DRL-based algorithm for trajectory tracking under wind disturbance. The agent learns to determine the rotation speed of each rotor of a hexacopter. A DDPG algorithm is used, but in addition to the existing DDPG algorithm, a policy relief (PR) method based on an epsilon-greedy exploration-based technique, and a significance weighting (SW) method are integrated into the DDPG framework. The former method improves the agent's exploration skills and its adaptation to environmental changes. The latter helps the agent update its parameters in a dynamic environment. In training, the implementation of PR and SW methods to the DDPG algorithm provides better exploration performance and faster convergence of the learning process, respectively, even in a dynamic environment. This method reaches a higher average reward and has a lower position error compared to the DDPG, DDPG with RP, and DDPG with SW. Also, this algorithm provides higher control accuracy compared to the cascaded active disturbance rejection control algorithm in terms of position, velocity, acceleration, and attitude errors.

Hwangbo et al.~\cite{hwangbo2017control} have proposed a method to increase UAV stabilization. A NN improves UAV stability training with RL. Monte Carlo samples are produced by on-policy trajectories and are used for the value function. The value network is used to guide policy training, and the policy network controls the quadrotor. Both are updated in every iteration. A new analytical measurement method describes the distance between action distribution and a new policy for policy optimization. This policy network gives an accurate reaction to step response. The policy stabilizes the quadrotor even under extreme situations. The algorithm shows better performance than DDPG and Trust Region Policy Optimization with a generalized advantage estimator (TRPO-gae) in terms of computation time.

Mitakidis et al.~\cite{mitakidis2023deep} have also studied the target tracking problem. A CNN-based target detection algorithm has been used on an octocopter platform to track a UGV. A DDPG-RL is applied in a hierarchical controller (instead of a position controller). The CNN learns to detect the UGV, and the DDPG-RL algorithm learns to determine the roll, pitch, and yaw actions in the outer loop of the controller. These actions are taken from the NN output as normalized between $[-1,1]$, but these normalized values are multiplied by a spectrum of acceptable values. While the roll and pitch actions expand between -3 and 3 degrees, the yaw action spreads between -5 and 5 degrees. An experiment is conducted with a low-altitude octocopter and with manual control of a UGV. Fluctuations are observed in the distance error due to the aggressive maneuver of the UGV, but overall, the results are good. 

Li et al.~\cite{li2020uav} have studied the target tracking problem in uncertain environments. The proposed approach consists of a TD3 algorithm and meta-learning. The used algorithm is named meta twin delay deep deterministic policy gradient (meta-TD3). TD3 learns to control the linear acceleration of the UAV and the angular velocity of the heading angle. The state space includes the position of the quadrotor in the x-y plane, the heading angle, the linear velocity, the angle between the motion direction and the straight line between the UAV and the target, and the Euclidean distance between the UAV and the target. Meta-learning overcomes the multi-task learning challenge. Tasks are trajectories of the ground vehicle that is followed. A reply buffer is built for the task experience. When the agent interacts with the environment, the state space, action space, reward value, and the next step state space that corresponds to the task are saved into the reply buffer. The method provides a significant improvement in convergence value and rate. Meta-TD3 adapts to the different movements of the ground vehicle faster than TD3 and DDPG algorithms. Meta-TD3 tracks the target more effectively.

Panetsos et al.~\cite{panetsos2022deep} have offered a solution for the payload transportation challenge using a DRL approach. An attitude PID controller is used in the inner loop of the cascaded controller structure, while a position controller in the outer loop is replaced with a TD3-based DRL algorithm. The DRL algorithm in the outer loop learns to create the reference Euler angles, roll and pitch, and the reference translational velocity of the octocopter on the z-axis. The method controls the system successfully to reach desired waypoints. 

Wang and Ye~\cite{wang2022consciousness} have developed a consciousness-driven reinforcement learning (CdRL) for trajectory tracking control. The CdRL learning mechanism consists of online attention learning and consciousness-driven actor-critic learning. The former selects the best action. The latter increases the learning efficiency based on the cooperation of all subliminal actors. Two different attention-learning methods are utilized for online attention learning: short-term attention learning and long-term attention learning. The aim of the former is to select the best action. The latter selects the best action to sustain the system's stability. The long- and short-term attention arrays are combined to make a decision about which actor should be given more attention. This learning algorithm is compared with Q-learning; the position error in the proposed algorithm is lower than in Q-learning. The same is also seen in the velocity error. However, this method is slightly better than Q-learning when it comes to attitude error. The UAV is successfully controlled to track the desired trajectory by the CdRL algorithm.

Xu et al.~\cite{xu2023omnidrones} have created a benchmark using PPO, SAC, DDPG, DQN algorithms for single-agent tasks and multi-agent PPO (MAPPO), heterogeneous-agent PPO (HAPPO), multi-agent DDPG (MADDPG), and QMIX algorithms for multi-agent tasks with different drone systems. Single-agent tasks include hovering, trajectory tracking, and flythrough. Multi-agent tasks cover hover, trajectory tracking, flythrough, and formation. To increase task variation, payload, inverse pendulum, and transportation challenges are integrated into the single- and multi-agent tasks. Learning performance differs based on specific tasks. 

\section{Online learning} \label{sec: online learning}

In online learning, an agent learns and it is updated during the system operation by incorporating collected data from sensors used to make and improve decisions, also interacting with the environment. The agent learns in real-time, enhancing its decision and prediction capabilities (with respect to the assigned mission). Before proceeding, it is important to mention three more perspectives that clarify further what online learning is (compared to the definition provided in Section~\ref{sec: background information}) and also they offer a more 'rounded' and more 'open-minded' point of view to consider. 

\text{Srinivasan and Jain~\cite{srinivasan2010innovations}}, state that the intelligent behavior of an agent may be limited given the extreme difficulty in developing knowledge structures and rule bases, which completely describe the task of the agent if the task by nature is complex. This problem can be partially overcome by causing the agent to learn on its own during this task. 

\text{Hoi et al.~\cite{hoi2021online}}, define online learning as a method of ML based on which data is arriving in sequential order, and where a learner aims to learn and to update the best prediction for future data at every step. Online learning is able to overcome the drawbacks of batch learning because the predictive model can be updated instantly for any new data instances.

\text{Otterlo and Wiering~\cite{van2012reinforcement}}, state that an important aspect during the learning task (during the learning process) is the distinction between online and offline learning. The difference between these two types is influenced by factors such as whether one wants to control a real-world entity or whether all necessary information is available. Online learning performs learning directly on the problem instance. Offline learning uses a simulator of the environment as a cheap way to get many training examples for safe and fast learning.

The literature review (from 2015) reveals that 11 papers have used online learning-based algorithms to control multirotor UAVs, in particular quadrotors. Their review and comparison provide a more accurate understanding of what the research is all about, what is being learned, why, and how, and what results have been produced. 

Yang et al.~\cite{yang2018leader} have developed optimal control protocols to solve the distributed output synchronization problem for leader-follower multiagent systems. The adopted RL algorithm solves the underlying non-homogeneous Algebraic Riccati Equations (AREs) in real-time - this is basically a distributed optimal tracking problem. Solving the AREs guarantees synchronization of the followers' and the leader's output. A distributed observer is derived to forecast the leader's state and to produce the reference signal for each follower. The proposed RL algorithm does not require knowledge of the quadrotor dynamics. This method gives better results than the adaptive control approach developed by Das and Lewis~\cite{das2010distributed}. Moreover, when the transient response of the output synchronization error for each follower is compared, the error in \cite{yang2018leader} converges to zero faster than in \cite{das2010distributed}.

\begin{figure}
\begin{center}
\includegraphics[width=10.5 cm]{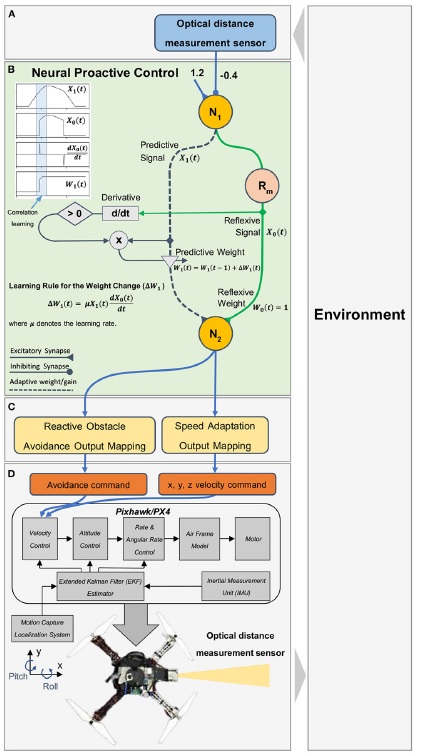}
\caption{Online learning controller scheme of \cite{jaiton2022neural}. \label{Three neurons}}
\end{center}
\end{figure}

A neural proactive closed-loop controller has been developed in ~\cite{jaiton2022neural} that learns control parameters within a few trials and in a computationally inexpensive way. No knowledge of the UAV dynamic model is required. This technique has been used for quadrotor speed adaptation and obstacle avoidance following the detailed block diagram configuration shown in Fig.~\ref{Three neurons}. For performance evaluation purposes, an MPC and the proposed method are implemented and compared. The MPC requires knowledge of the quadrotor dynamic model. Thus, the neural proactive controller method is more appealing. It also provides robust results for speed adaptation even in the presence of wind disturbances. This approach has been implemented and tested on UAVs with different dynamics, in a Gazebo environment, under different maximum flying speeds. The neural proactive controller generates a control command 56.32\% faster (on average) than the MPC; it is also 99.47\% faster than the MPC in total learning and optimization time. The UAV is trained within 3-4 trials adapting its speed and learns to stay away from obstacles at a safe distance.  

Shin et al.~\cite{shin2020evolution} have applied online learning to optimize speed parameters with the aim to complete  missions in less time (i.e., drone races through gates). A pre-trained actor network, modified to be suitable for online training, detects a specific object. The control network, which comprises the online learning component, provides UAV optimized linear velocity and acceleration in the x, y, and z directions, as well as the distance to a gate to pass through with maximum velocity, and without collisions. Each gate and the gate center are detected by a monocular camera system. The UAV coordinates are determined based on the provided information, and a loss function is used to make sure that the quadrotor passes through the center of the gate. The maximum velocity, acceleration, and threshold distance (that determines whether or not the quadrotor reaches the gate location) are parameters that are optimized for each gate. They are updated each time the quadrotor passes through each gate. As opposed to several conventional navigation approaches, this method requires a single neural network to control the quadrotor. The derived network has successfully handled all race processes in environments with demanding obstacle avoidance and navigation requirements. When tested in real-time, the quadrotor completed the race track in around 180 seconds in the first time, but in about 60 seconds after about 9000 times.

Sarabakha and Kayacan~\cite{sarabakha2019online} have proposed an online learning-based control method to improve UAV trajectory tracking. Total thrust and the three torques around the x, y, and z axes are used as control inputs. The online learning component learns to update the weight of a DNN to improve control performance. This method consists of two phases, pre-training called offline learning, and, post-training called online learning. In pre-training, supervised learning is used to control the quadrotor by mimicking a PID controller (using an input-output dataset for a set of trajectories). When the quadrotor is controlled by the trained DNN controller, a fuzzy logic system (FLS) keeps training the DNN online by providing feedback about its performance. The offline learning based performance is not different from the classic PID controller performance. However, online learning allows for the system to accurately predict evolution and desired signal estimation. This approach has been tested in slow circular, fast circular, and square-shaped trajectories. Performance of the well-tuned PID controller used in offline training, offline-trained DNN, and the online training approach are compared, with the last one clearly showing better performance for the slow and fast circular trajectories. The square-shaped trajectory was not used in the offline training phase. When the three are implemented and tested in the square-shaped trajectory, again, the online learning method slightly outperforms other approaches.

O'Connell et al.~\cite{o2023pretraining} have proposed Neural-Fly that includes an online learning phase to overcome instabilities caused by wind effects. The approach includes offline and online learning. In offline learning, the DNN output is a set of basis functions that represent aerodynamic effects. The latter phase is an online adaptive control phase that learns to adapt the control system to new wind conditions rapidly and robustly. For offline learning, the domain adversarially invariant meta-learning (DAIML) algorithm is developed to learn aerodynamic effects under wind-invariant conditions. A stochastic gradient descent method is used in the DAIML algorithm for training. For online learning, a Kalman Filter-based adaptation algorithm estimates the wind-dependent linear coefficients. The position tracking error and the aerodynamic force prediction error terms are utilized in this estimation under wind-variant conditions. The online learning component provides fast adaptation to wind variations. This approach may be used to control several quadrotors without the need to pre-train each UAV. Neural-Fly shows better performance than the nonlinear tracking controller found in \cite{mellinger2011minimum}, $\mathcal{L}_1$ adaptive controller, and an incremental nonlinear dynamics inversion controller when the quadrotor is subjected to time-variant wind profiles. 

Jia et al~\cite{jia2023research} have provided a solution to the trajectory tracking problem by combining a fuzzy logic method, a radial basis function (RBF) NN, and a classical PID controller. The PID output and the current UAV position information are provided to the RBF NN as input values, and the network learns to adjust controller parameters. The fuzzy logic component selects the initial controller parameters, while the RBF NN adjusts them. Both are combined to create a new set of parameters for the PID controller. In the fuzzy logic component, a database created from expert knowledge is used to decide about the initial controller parameters. However, this dataset cannot be used in the RBF NN component since the adopted algorithm is an unsupervised learning method. However, the combination of both methods overcomes this limitation and provides online learning abilities. When the fuzzy logic system adjusts the PID gains, the RBF NN tweaks the PID parameters to overcome PID weaknesses caused by environmental disturbances. This approach when compared with PID and fuzzy-PID (FPID) controllers shows better performance for trajectory tracking.  

Zhang et al.\cite{zhang2023realtime} have investigated the problem of adaptive control and have offered a real-time brain-inspired learning control (RBiLC) method as a solution. Three attitude angle errors are set as input and a NN provides the control parameter increment as output. In the RBiLC method, a PID controller is used, and the controller parameter is updated in each interaction. A DRL method learns the controller parameter rate. The algorithm uses the Nesterov momentum technique for gradient descent. Controller stability and convergence of the tracking error is demonstrated. The quadrotor takes off and reaches a 10-meter altitude, then it hovers for 3 seconds in this position with the initial controller parameters. The learning algorithm is activated using a switch system, and the RBiLC algorithm updates PID gains in real-time in an environment with wind disturbances. This approach learns controller parameters in 3 to 5 minutes, which is a much shorter time than in offline learning methods. This method shows significant improvement for stabilization in roll and pitch angles, but performance is not the same in the yaw angle. Under wind disturbances, the method provides a shorter rise time and steady-state time for roll and pitch when compared to the classic PID controller.  

Shiri et al.~\cite{shiri2020remote} have studied the online path planning problem using NN-based opportunistic control. An online-trained NN learns to solve the Hamilton-Jacobian-Belmann (HJB) equation in real-time. The opportunistic HJB, oHJB,  control algorithm learns whether it will upload the control action (aHJB) that is the output of the NN or the current online trained NN (mHJB). A base station (BS) is utilized to handle the online learning algorithm. The UAV state is downloaded to the BS. First, the NN is trained online to solve the HJB in real-time, in the BS. Then, the output of the trained NN (the control action aHJB), is uploaded to the UAV. Secondly, the current online trained NN (represented as mHJB), is uploaded to the UAV (instead of the aHJB), and the mHJB is fed with the current states. Then, the UAV takes the action that is locally assessed by the uploaded mHJB in real-time. The oHJB control provides the decision mechanism to switch from aHJB to mHJB according to the connection between the multirotor UAV and the BS. Based on the oHJB, the UAV can keep taking actions using the last uploaded mHJB even if it loses connection with the BS. Since the size of the trained NN model in the BS is larger than the size of the action space, a trade-off occurs between uploading delays and control robustness against poor channel conditions. The oHJB arrives at the targeted location in a shorter time than aHJB and mHJB. The aHJB and mHJB may fail to arrive at the desired location.

Wang et al.~\cite{wang2018safe} have used a data-driven approach (Gaussian processes) to learn quadrotor models applied in partially unknown environments. Barrier certificates are learned and utilized to adjust the controller to not encroach in an unsafe region. A safe region is certified and it is progressively spread with new data. Collecting more data points about system dynamics reduces uncertainty and maximizes the volume of the safe region. Discretizing the state space helps sample a finite number of points to affirm the barrier certificates; so adaptive sampling decreases the number of the required sampling points. Also, the state space is adaptively sampled by enhancing the Lipschitz continuity of the barrier certificates without taking any risk on safety guarantees. Sampling the most uncertain state in the safe set of the system boosts learning efficiency during the exploration phase. A kernel function is utilized to decide data relevance, and 300 data points are chosen in the recursive Gaussian process by eliminating irrelevant data points online. The approach reduces the possibility of an accident occurring during the online learning phase. The adaptive sampling strategy provides significantly better results in decreasing the required sample points. The quadrotor never violates safe and unsafe regions; it uses barrier certificates to regulate the learning algorithms. A high probability of safety guarantee is produced by a Gaussian process for the dynamic system, and this process learns unmodeled dynamics to help the quadrotor successfully stay within the safe region. When the tracking error of the learning-based controller is compared to the tracking error without Gaussian process inference, the tracking error is much smaller in the learning-based controller. 

He et al.~\cite{he2019state} have solved the time-varying channel(s) problem on air-to-ground links caused by UAV mobility in dynamic environments. Offline learning learns to minimize the prediction loss function in the prediction network and the evaluation loss function in the evaluation network. In online learning, the aim is to learn to minimize the difference between two networks. A state-optimized rate adaptation algorithm called StateRate is developed to solve this problem using the onboard UAV sensors. The evaluation and prediction networks are exploited in online training. The received signal strength indicator and the channel state information are used as inputs for both networks; the prediction network, additionally, uses the UAV's states as input. The output of the evaluation NN is used for supervision and compared with the output of the prediction network by using a fully connected layer. The StateRate algorithm accurately predicts the optimal rate. The rate prediction is handled as a multi-class classification problem using online learning. This method is applied and tested in a commercial quadrotor (DJI M100), and has shown better performance than the best-known rate adaptation algorithms applied in UAVs with 2-6 $m/s$ velocity.

\begin{table*}[]
\centering
\caption{Online Learning Papers}
\label{Tab: Online learning}
\resizebox{\textwidth}{!}{
\begin{tabular}{|c|c|c|c|c|c|c|c|c|}
\hline
\textbf{Year} & \textbf{Paper}      & \textbf{Task}     & \textbf{Algorithm}    & \textbf{Model-free or -based} & \textbf{Advantages}   & \textbf{Compared with}    & \textbf{Offline part} & \textbf{Sim/Exp} \\ \hline \hline
2018          & Yang et al. \cite{yang2018leader}                     &  \begin{tabular}[c]{@{}c@{}}Navigation\\ Synchronization \end{tabular}     & \cite{yang2018leader}                                                                    & Model-free                    & \begin{tabular}[c]{@{}c@{}}Solves inhomogeneous \\ algebraic Riccati equations online\end{tabular}                                                       & \begin{tabular}[c]{@{}c@{}}Adaptive control approach\\ in \cite{das2010distributed}\end{tabular}                                                                                     & No                    & Sim              \\ \hline
2018          & Wang et al. \cite{wang2018safe}                       & Environment Exploration                                                          & \begin{tabular}[c]{@{}c@{}}Data-driven approach\\ based on Gaussian Process\end{tabular} & Model-free                    & \begin{tabular}[c]{@{}c@{}}Reduces the possible crashes \\ in the online learning phase\end{tabular}                                                    & -                                                                                                                                                                                    & No                    & Sim              \\ \hline
2019          & Sarabakha and Kayacan \cite{sarabakha2019online}           & Trajectory tracking                                                              & Back-propagation                                                                         & Model-free                    & -                                                                                                                                                       & \begin{tabular}[c]{@{}c@{}}Offline trained network\\ PID controller\end{tabular}                                                                                                     & Yes                   & Sim              \\ \hline
2019          & He et al. \cite{he2019state}                          & \begin{tabular}[c]{@{}c@{}}Agile mobility in \\ dynamic environment\end{tabular} & StateRate                                                                                & Model-free                    & \begin{tabular}[c]{@{}c@{}}Adjusts finely the prediction framework, and \\ onboard sensor data are effectively used\end{tabular}                        & \begin{tabular}[c]{@{}c@{}} Previous OPT\\ Signal-to-noise rate (SNR) \\ SampleRate \\ CHARM \end{tabular}                                                                                                                                                          & Yes                   & Sim              \\ \hline
2019          & Wang et al. \cite{wang2019deterministic}              & Robust Control                                                                   & DPG-IC                                                                                   & Model-free                    & Elimination of the steady error                                                                                                               & \begin{tabular}[c]{@{}c@{}}PID controller\\ DDPG\end{tabular}                                                                                                                        & Yes                   & Sim              \\ \hline
 2020          & Shin et al. \cite{shin2020evolution}                  & Speed Optimization                                                               & SSD MobileNet                                                                            & Model-free                    & Quicker object detection time                                                                                                                           & -                                                                                                                                                                                    & No                    & Sim              \\ \hline
2020          & Shiri et al. \cite{shiri2020remote}                   & Path Planning                                                                    & oHJB                                                                                     & Model-free                    & \begin{tabular}[c]{@{}c@{}}The algorithm keeps working \\ even if UAV loses the connection with BS\end{tabular}                                         & \begin{tabular}[c]{@{}c@{}}aHJB\\ mHJB\end{tabular}                                                                                                                                  & No                    & Sim              \\ \hline
2022          & Jaiton et al. \cite{jaiton2022neural}                 & Speed optimization                                                               & Neural Proactive Control                                                                 & Model-free                    & Computationally inexpensive                                                                                                                             & MPC                                                                                                                                                                                  & No                    & Exp              \\ \hline
2023    & O'Connell et al.~\cite{o2023pretraining} & Stabilization  & DAIML & Model-free & \begin{tabular}[c]{@{}c@{}} Can control a wide range of quadrotors \\ Not require pretraining for each UAV\end{tabular} &  \begin{tabular}[c]{@{}c@{}} Mellinger \& Kumar~\cite{mellinger2011minimum} \\ $\mathcal{L}_1$ Adaptive Controller \\ Incremental Nonlinear Dynamics Inversion Controller \end{tabular}    & Yes   & Exp         \\ \hline

2023    & Jia et al.~\cite{jia2023research} & Trajectory tracking   & RFPID & Model-based   & Strong learning ability   & \begin{tabular}[c]{@{}c@{}} PID \\ Fuzzy-PID\end{tabular} & No    & Sim             \\ \hline

2023    & Zhang et al.~\cite{zhang2023realtime} & Stabilization   & RBiLC & Model-free    & \begin{tabular}[c]{@{}c@{}}Significant improvement for\\the stabilization in roll and pitch \\ Not show the same performance on yaw\end{tabular}    & PID   & No   &   Exp          \\ \hline
\end{tabular}%
}
\end{table*}

Wang et al.~\cite{wang2019deterministic} have worked on robust control of a quadrotor using a DRL-based controller. The method includes both offline and online learning. During offline learning, an offline learning control policy is learned. The actor-network learns the normalized thrusts produced by each rotor between 0 and 1. During online learning, the offline control policy continues to be optimized. In the offline phase, the Deterministic Policy Gradient-Integral Compensator (DPG-IC) algorithm is trained with random actor and critic NN weights in episodic structure. In the online learning phase, the trained DPG-IC is used as the initial NN structure that is trained continuously. The offline DPG-IC also runs alongside the online algorithm. The system switches to the offline algorithm when the states are close to the safety limits, and the quadrotor goes back to the safe range. The aim of the online learning phase is to close the gap between the simplified model and the model with real flight dynamics. DPG-IG eliminates the steady-state error. A well-tuned PID controller shows similar performance compared to the offline DPG-IC policy when the size of the quadrotor is increased from 0.12 m to 0.4 m, but for larger sizes, the PID controller is not stable while the proposed method shows successful performance for sizes up to  1 m. The PID controller and the offline DPG-IC algorithm are also compared for different payloads. The payload of the quadrotor increases by 10\% each step; the PID cannot control the quadrotor with a 30\% increased payload. On the other hand, the proposed offline learning method successfully controls the quadrotor with up to 50\% payload increase. Then, the performance of the online learning policy is compared with the offline learning policy. The model with 0.4 m radius and 20\% increased payload is used. After only 200 online training steps, performance is significantly increased. 

In summary, since RL is built on the interaction between the agent and environment and it is based on a reward system, RL algorithms are useful for online learning and they are widely preferred for applications related to control of multirotor UAVs. 

\section{Discussion and Conclusions} \label{sec: discussion and conclusion}

This survey provides summary results that are combined in four Tables. The main two tables reflect offline and online learning techniques and algorithms. Both tables offer a clear picture of the publication year, the adopted method, the task/mission, and also what is being learned. Coupled with the provided information for each method, the reader acquires information about the applicability and implementability of each technique, as well as about the specific application the approach has been developed for. The other two tables offer specific details 
on RL approaches and on the value function-based and policy search-based methods. Overall, this survey provides a comprehensive overview of the evolving landscape of multirotor UAV navigation and control, particularly focusing on the integration of learning-based algorithms over the past decade. 

Given that eventually, there will be 'almost infinite computational power' that will require 'almost zero computational time' to return results, this survey paper offers a starting point for subsequent studies on what is hard real-time implementable.

\section*{Acknowledgments}
This work is part of Serhat S\"onmez's PhD dissertation research. Serhat S\"onmez is the main contributor in this paper. This research was partially supported by the Ministry of National Education of the Republic of Turkey on behalf of Istanbul Medeniyet University and the D. F. Ritchie School of Engineering and Computer Science, University of Denver, CO 80208.

\end{document}